\documentclass[runningheads]{llncs}
\usepackage{graphicx}

\usepackage{tikz}
\usepackage{comment}
\usepackage{amsmath,amssymb} 
\usepackage{color}
\usepackage[width=122mm,left=12mm,paperwidth=146mm,height=193mm,top=12mm,paperheight=217mm]{geometry}
\usepackage{graphicx}
\usepackage{booktabs}
\usepackage[pagebackref,breaklinks,colorlinks]{hyperref}
\usepackage{hyperref}

\usepackage[accsupp]{axessibility}  

\begin{document}
\pagestyle{headings}
\mainmatter
\def\ECCVSubNumber{2153}  

\title{Perceptual Artifacts Localization for Inpainting} 


\titlerunning{Perceptual Artifacts Localization for Inpainting}
%
\author{Lingzhi Zhang\inst{1} \and
Yuqian Zhou\inst{2} \and Connelly Barnes\inst{2} \and Sohrab Amirghodsi\inst{2} \and \\ Zhe Lin\inst{2} \and Eli Shechtman\inst{2} \and Jianbo Shi\inst{1}}

\authorrunning{Zhang et al.}
%
\institute{University of Pennsylvania \and
Adobe Research \\
}
\maketitle
\begin{figure*}[!h]
    \centering
    \vspace{-26 pt}
    \includegraphics[trim=0.0in 2.8in 1.7in 0in, clip,width=\textwidth]{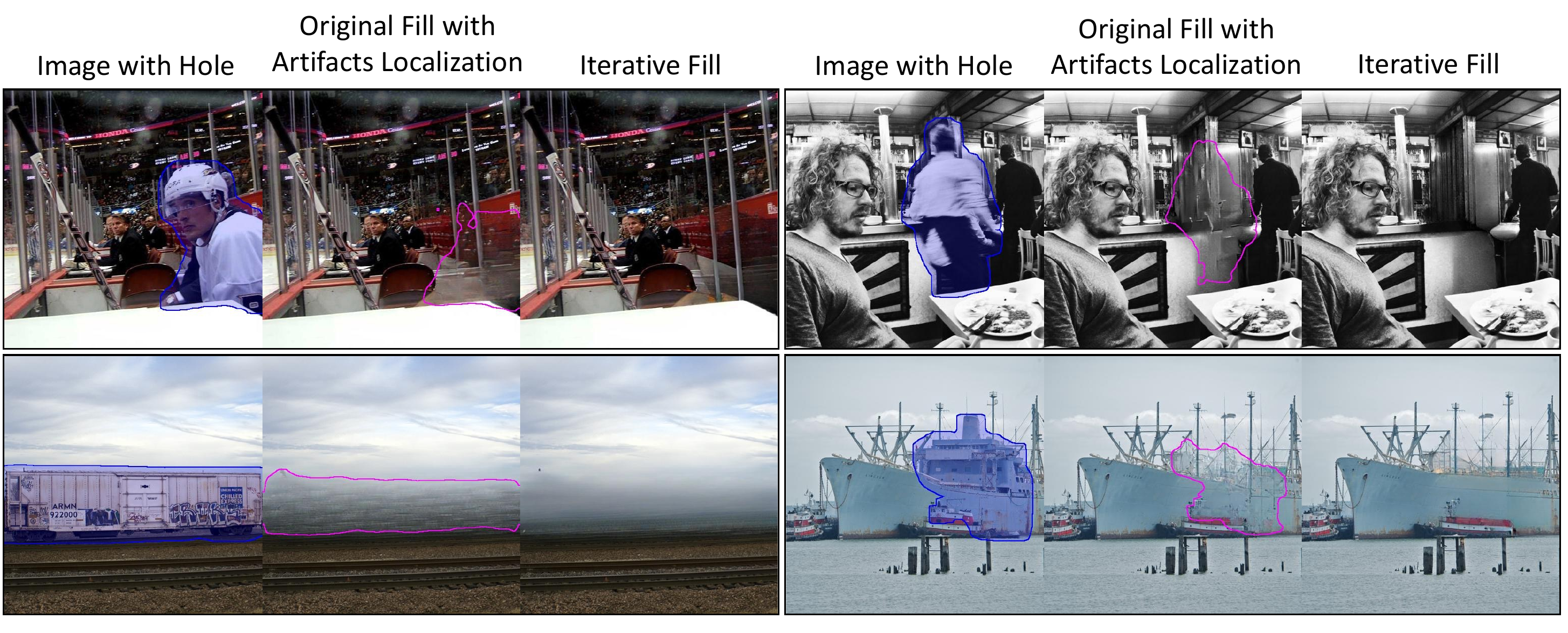}
    \vspace{-25 pt}
    \caption{Visual examples to show that our segmentation network can reliably localize the perceptual artifacts region, as indicated by the pink boundary in the second columns. Given the artifacts localization, we enable the inpainting model \cite{su2020blindly} iteratively fill on the artifacts region to obtain better inpainting quality, as shown in the third columns. }
    \label{fig:teaser}
    \vspace{-30 pt}
\end{figure*}

\begin{abstract}
Image inpainting is an essential task for multiple practical applications like object removal and image editing. Deep GAN-based models greatly improve the inpainting performance in structures and textures within the hole, but might also generate unexpected artifacts like broken structures or color blobs. Users perceive these artifacts to judge the effectiveness of inpainting models, and retouch these imperfect areas to inpaint again in a typical retouching workflow. Inspired by this workflow, we propose a new learning task of automatic segmentation of inpainting perceptual artifacts, and apply the model for inpainting model evaluation and iterative refinement. Specifically, we first construct a new inpainting artifacts dataset by manually annotating perceptual artifacts in the results of state-of-the-art inpainting models. Then we train advanced segmentation networks on this dataset to reliably localize inpainting artifacts within inpainted images. Second, we propose a new interpretable evaluation metric called  Perceptual  Artifact  Ratio  (PAR),  which is the ratio of objectionable inpainted regions to the entire inpainted area. PAR demonstrates a strong correlation with real user preference. Finally, we further apply the generated masks for iterative image inpainting by combining our approach with multiple recent inpainting methods. Extensive experiments demonstrate the consistent decrease of artifact regions and inpainting quality improvement across the different methods. Dataset and code are available at: {\href{https://github.com/owenzlz/PAL4Inpaint}{https://github.com/owenzlz/PAL4Inpaint}}

\end{abstract}


\section{Introduction}
\vspace{-5 pt}

Deep GAN-based image synthesis methods have been continuously improving image inpainting performance \cite{yu2021diverse,zhao2020uctgan,liu2021pd,cai2020piigan,peng2021generating,zhang2020multimodal,zeng2020high,zhao2021large,suvorov2021resolution}
for practical applications like object removal and image editing. Due to the ill-posed nature of image inpainting tasks, when encountering large holes or complex structures \cite{suvorov2021resolution} within the hole, image inpainting becomes extremely challenging. Along with almost all state-of-the-art algorithms, inpainting artifacts tend to appear in the generated images. Those artifacts mostly include broken structures or color bleeding in the traditional patch synthesis methods \cite{barnes2009patchmatch}, imperfect structures like disconnected or distorted lines, GAN-based generation artifacts or color blobs.  In typical retouching workflows, users tend to judge the inpainting performance by those artifacts, and fix them by drawing masks on those regions and re-runing the automatic inpainting tools. Therefore, localizing and segmenting those artifacts is intuitively and naturally beneficial for inpainting algorithm evaluation and performance improvement.


Intuitively, finding more or larger artifacts within the hole area indicates a worse inpainting performance. Traditionally, image inpainting is regarded as an image reconstruction and restoration problem, and commonly-used metrics like PSNR, MSE and LPIPS \cite{zhang2018unreasonable} are utilized to compare the inpainted result to the original image in terms of content or pixel similarity. However, in many cases, image inpainting is used for foreground object removal \cite{criminisi2003object,criminisi2004region}. Users prefer a visually plausible background generation rather than a faithful foreground reconstruction. Other quantitative metrics like Frechet Incept Distance (FID) \cite{heusel2017gans,parmar2021buggy} and Paired/Unpaired Inception Discriminative Score (P/U-IDS) \cite{zhao2021large} are computed on the entire images over large evaluation datasets. We are lacking in an intuitive metric which is more interpretable, operates on localized hole regions, and supports single result evaluation. Therefore, an automatic and reliable artifacts segmentation network may fill the gap. 

In practical inpainting applications, users may choose to manually fix those artifacts by re-masking perceptually bad regions and re-running the models. Intuitively, after a couple of iterations, inpainting results are expected to be largely improved compared with the initial ones. Iterative hole filling has been studied in deep learning pipelines \cite{zeng2020high,oh2019onion,guo2019progressive}, and is shown to outperform one-pass inpainting. But the masks used in each iteration are either unreliable ones \cite{zeng2020high} learned with image reconstruction loss or predefined eroded masks \cite{oh2019onion,guo2019progressive}. Hence, an automatic artifacts segmentation network can effectively detect the perceptual artifacts in each iteration, and make the iterative filling run in a more efficient and effective way.

Although these inpainting artifacts are easily identifiable by humans, very few studies \cite{zeng2020high} have developed models to automatically detect and localize these artifacts in inpainting results. Researchers have studied identifying manipulated or synthesized images \cite{wang2020cnn,cozzolino2018forensictransfer,wang2019fakespotter,zhang2019detecting,marra2018detection,wang2020cnn,chai2020makes}, edited image regions \cite{chai2020makes}, or the entire inpainted image regions \cite{li2019localization,wu2021giid,li2021noise}. However, automatic localization of those artifacts within the inpainted holes was seldom discussed. This is mainly because a representative and well-organized dataset consisting of image inpainting results and artifact annotations is not yet available. Using the knowledge and expertise of professional photographers, deep networks can learn to efficiently detect and segment these artifacts.

In this paper, inspired by a typical user workflow when using inpainting tools, we assume that automatic perceptual artifacts segmentation for image inpainting will potentially benefit algorithm evaluation and boost inpainting performance. To verify our hypothesis, we collect inpainting results generated by multiple state-of-the-art deep inpainting models and annotate pixel-wise artifacts with a team of human professionals, and benchmark the dataset using advanced segmentation networks. Our proposed artifacts localization network outputs a binary mask highlighting the artifacts region. This mask can be used to: (1) compute the occupation ratio over the hole mask to evaluate and compare different inpainting algorithms on single test image without ground truth, and (2) achieve iterative filling to progressively improve inpainting performance. In summary, our contributions are in three folds:

\begin{itemize}
    \item We study the importance of a novel task, inpainting artifacts segmentation. Given its strengths in inpainting evaluation and result refinement, we construct a dataset consisting of 4,795 inpainting results with per-pixel perceptual artifacts annotations. We further benchmark the dataset using multiple segmentation network structures and analyze the human subjective factors in detail. Extensive experiments demonstrate its robustness on state-of-the-art inpainting models. 
    
    \item We present the Perceptual Artifact Ratio (PAR) calculated from the artifact area detected inside the hole. PAR is an interpretable, intuitive, simple yet effective evaluation metric for comparing inpainting algorithms on a single image without ground truth. Our metric makes it possible to automatically evaluate object removal performance. Our user study also shows that PAR correlates more strongly with real user preferences than other metrics. 
    
    \item We applied the artifacts segmentation network to iterative filling pipeline. After each iteration, we visualize that the detected artifact regions are consistently shrinking for all the tested inpainting models, and the results are refined with better structures and colors. Another user study suggests that iterative filling using our proposed artifacts masks will likely not degrade the inpainting performance and in many cases improve it. 
\end{itemize}

\vspace{-10pt}
\section{Related Work}
\vspace{-5pt}

\subsection{Image Inpainting}
\vspace{-5pt}

Classical image inpainting methods include diffusion-based methods \cite{bertalmio2000image,ballester2001filling} that propagate information from the boundary inwards to fill the hole, and patch-based methods \cite{barnes2009patchmatch} that search for the reference region to fill the hole. On the rise of deep learning, researchers proposed deep models to improve the inpainting performance from diverse angles, such as attention mechanism \cite{iizuka2017globally}\cite{yu2018generative}\cite{liu2019coherent}\cite{xie2019image}\cite{liu2018image}\cite{yu2019free}\cite{wan2021high} \cite{suvorov2021resolution}, loss function and discriminator design \cite{iizuka2017globally}\cite{yang2020learning}\cite{yeh2017semantic}\cite{yu2018generative}\cite{zeng2019learning}, progressive \cite{guo2019progressive}\cite{zeng2020high} \cite{li2020recurrent} \cite{li2019progressive}\cite{li2019progressive}\cite{zhang2018semantic} or multiscale \cite{yang2017high}\cite{yi2020contextual}\cite{zeng2019learning}\cite{wang2018image} architectures, use of intermediate guide respresentation \cite{yu2018generative}\cite{nazeri2019edgeconnect}\cite{ren2019structureflow}\cite{song2018spg}\cite{liao2020guidance}\cite{xiong2019foreground}\cite{wang2020multistage}\cite{wang2021image}\cite{guo2021image}, and multimodal plausible outputs \cite{zheng2019pluralistic}\cite{yu2021diverse}\cite{zhao2020uctgan}\cite{han2019finet}\cite{liu2021pd}\cite{cai2020piigan}\cite{peng2021generating}\cite{zhang2020multimodal}. Among these works, ProFill \cite{zeng2020high}, CoMod-GAN \cite{zhao2021large}, and LaMa \cite{suvorov2021resolution} are the most recent leading models. ProFill \cite{zeng2020high} proposed to implicitly learn a confidence map that guides the generator to iteratively fill the hole, as well as a attention-guided refinement module to upsample the output. CoMod-GAN \cite{zhao2021large} leveraged the StyleGAN architecture \cite{karras2020analyzing} to conditionally synthesize filled region, where their filled content could be creative and not necessarily existed in the context. Finally, LaMa \cite{su2020blindly} integrated the fast Fourier convolution \cite{chi2020fast} to effectively capture global contextual information, and set the new state-of-the-arts. The goal of our work is to detect and localize the perceptual artifacts in the filled images independent of the inpainting models, and thus is in an orthogonal direction to these previous inpainting works.

\vspace{-10pt}
\subsection{Image Inpainting Quality Assessment}


There are two types of commonly used metrics for image inpainting. The first quantifies the performance for a whole dataset of generated images. These metrics include Frechet Incept Distance (FID) \cite{heusel2017gans}\cite{parmar2021buggy} and Paired/Unpaired Inception Discriminative Score (P/U-IDS) \cite{zhao2021large}, which measure the distance between the distribution of generated and real images using the deep Inception features \cite{szegedy2016rethinking}. For single image quality assessment, previous works often 
treat inpainting as a reconstruction task and thus compare the filled image with the original image using the reconstruction metrics, such as MSE, SSIM, PSNR or LPIPS \cite{zhang2018unreasonable}. This is reasonable only when holes are sampled on the background region. When holes largely overlap with or totally cover a foreground object, the current models would mostly fill the hole using background pixels, which is totally irrelevant to the original content. In these cases, the filled region of object removal could look natural and realistic, but could be totally different from the original object. Thus, reconstruction metric would no longer be a proper metric. Other potential assessments for measuring single image inpainting quality of object removal are No-Reference Image Quality Assessment (NR-IQA) methods \cite{fang2020perceptual}\cite{su2020blindly}\cite{talebi2018nima}\cite{ying2020patches}\cite{zhang2018blind}\cite{ke2021musiq}. Although previous inpainting works have rarely used NR-IQA metrics, we tried out two recent methods Hyper-IQA \cite{su2020blindly} and MUISQ \cite{ke2021musiq}, and found that MUISQ \cite{ke2021musiq} has a relatively reasonable correlation with human perception compared to the other existing metrics for measureing object removal inpainting quality. In this work, we aim to use the area size of localized inpainting artifacts as a no-reference metric to measure the quality of hard case inpainting, which is object removal. Experimental studies in section 5 show that our proposed metric outperforms both reconstruction-based metrics and existing NR-IQA in terms of correlation with human perception.

\vspace{-10pt}
\subsection{Detecting Artifacts in Generated Images}

Other related works include detecting the generated/fake images and localizing the manipulated region in the image. One line of works \cite{cozzolino2018forensictransfer}\cite{wang2019fakespotter}\cite{zhang2019detecting}\cite{marra2018detection}\cite{wang2020cnn}\cite{chai2020makes} have studied training a binary classifier to classify the generated images, and Wang et al. \cite{wang2020cnn} has shown surprising generalization on diverse and unseen model outputs by detecting the common artifacts in CNNs/GANs. Chai et al. \cite{chai2020makes} proposed patch-based classifier to localize the region that causes the fake image detectable. Another line of works \cite{cozzolino2015splicebuster}\cite{huh2018fighting}\cite{rao2016deep}\cite{wang2019detecting} have proposed techniques to detect general image manipulations, such JPEG and resampling, other than GANs. In the image inpainting domain, Li et al. \cite{li2019localization} first proposed to use the high-pass filter CNNs to detect the inpainting region given the filled image. Later, Wu et al. \cite{wu2021giid} and Li et al. \cite{li2021noise} further improves the generalization of the mask detection to diverse inpainting models by proposing novel architecture or explicitly process high-frequency noise residual. Although all these works are related to us, a fundamental difference is that we aim to detect the perceptual artifacts that are judged by humans rather than simply detecting high-frequency noise/artifacts in the generated images. More specifically, in the inpainting context, our system detects the perceptual artifact region rather than the whole mask region, where perceptual artifact region is often a small subset of the mask. Thus, our work is essentially a different task from \cite{li2019localization}\cite{wu2021giid}\cite{li2021noise}.

\vspace{-10pt}
\section{Dataset Labeling and Statistics}
\label{data_labeling}
\vspace{-5pt}

In order to train a system that can detect the perceptual artifacts in the inpainted images, we build a dataset that consists of 4,795 images with per-pixel perceptual artifacts labels from humans. We use three leading inpainting models ProFill \cite{zeng2020high}, CoMod-GAN \cite{zhao2021large}, and LaMa \cite{suvorov2021resolution} to generate images to label. A labeling interface and a few examples of the labeled images are shown in the left and right of Fig.\ref{fig:data_labeling}, respectively. During labeling, we provide the users a filled image without showing the original image, and ask users to label regions with perceptual artifacts on their tablets. We intentionally do not include the original images in the interface, since otherwise users might have bias to compare everything with the original content in the hole. As we have discussed in section 2.2, the filled image could look natural and realistic, even though it's very different from the original image. We also put dilated bounding box around the hole region to help users more easily find the labeling region and focus on it. We intentionally do not indicate the hole mask in the image, so that the workers do not have any bias labeling around the hole boundary, and thus can purely make judgement based on the perceptual quality. In this case, since workers do not know where the hole is, their labeling might go over the hole boundary. However, this is not an issue, as we simply intersect the hole masks with the human labels as a  post-process. In addition, we provide a duplicate of the filled image in the interface, so that when workers can see the unbrushed filled image as reference during brushing. 

\begin{figure*}[t]
    \centering
    \includegraphics[trim=1.9in 3.4in 2.0in 0in, clip,width=\textwidth]{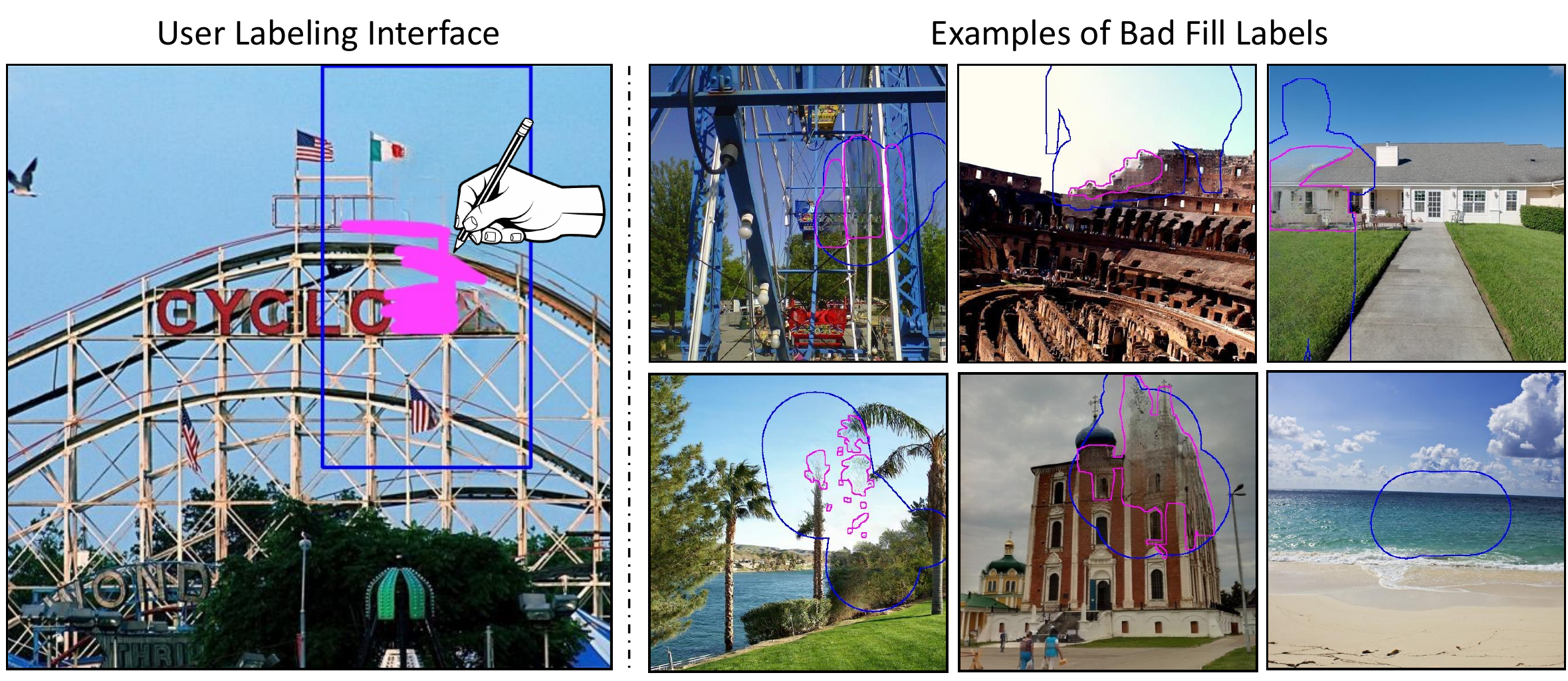}
    \vspace{-20 pt}
    \caption{The left is an illustration of interface that users label on the inpainted image, where the blue bounding box indicates the region the users should inspect. On the right, we show a few examples of bad fill labels from users, where the blue and pink boundaries indicate the holes and user labels, respectively. }
    \label{fig:data_labeling}
    \vspace{-10 pt}
\end{figure*}

From the labeling perspective, one fundamental challenge is that this task is highly subjective compared to traditional segmentation, since different people may have different opinions or standards to judge perceptual artifacts. Therefore, in order to standardize the labeling as much as possible, we recruit and train a professional team for this task. We use two rounds of checks to avoid "missed labeled" or "overly labeled" regions. In the first round, the professional workers cross check the results with each other. In the second round, a single human expert checks through all the labels. On average, approximately 10$\%$ of labels have been rectified during the checking process. In addition, in order to generate high-quality labels, we recruited five human experts with photography or design background in our team to label these images. Since these workers are heavy users of image editing tools, i.e. Photoshop, their labeling criterion could better reflect the common unsatisfactory/retouch regions in the hole filling process. 

Among these 4,795 images, there are 832 images that have nearly perfect fills, and thus workers did not label anything on these images. Although these images do not have segmentation labels, adding them into training could effectively help the network avoid predicting false positives. In terms of size of the labeling region, we found that the averaged ratio of "perceptual artifacts region / hole mask region" is 29.67\%. This number once again shows that detecting perceptual artifacts is fundamentally different from detecting the hole mask in \cite{li2019localization}\cite{wu2021giid}\cite{li2021noise}. We plan to release our dataset to the community for future research.


\vspace{-5pt}
\section{Perceptual Artifacts Segmentation}
\vspace{-5pt}

In this section, we discuss the details of our segmentation model along with extensive ablation studies. During training, we used "8:1:1" ratio to randomly split the train/val/test set. In total, we have 3,836 training images, 480 validation images, and 479 test images. In each model training, we use the validation set to select the best checkpoint, and evaluate the performance on the test set. All of our models are trained and evaluated using the MMSegmentation codebase \footnote[1]{MMSegmentation github: \href{https://github.com/open-mmlab/mmsegmentation}{https://github.com/open-mmlab/mmsegmentation}}.


\vspace{-5pt}
\subsection{Ablation Studies}
\vspace{-5pt}

\begin{table*}[t!]
    \begin{center}
     \resizebox{\textwidth}{!}{
      \begin{tabular}
      {c|c|c|c|c}
        \toprule 
        \textbf{Models} & \ \hspace{10pt} IoU \hspace{10pt} & \ \hspace{0pt} Precision \hspace{0pt} & \ \hspace{3pt} Recall \hspace{3pt} & \ \hspace{3pt} Fscore \hspace{3pt} \\
        \midrule 
        ResNet-50 backbone \cite{he2016deep} + HRNet head \cite{wang2020deep} & 41.35 & 58.45 & 58.56 & 58.51   \\
        \midrule
        Swin-B backbone \cite{liu2021swin} + Uper head \cite{xiao2018unified} & 44.20 & 63.01 & 59.69 & 61.30   \\
        \midrule
        ResNet-50 backbone \cite{he2016deep} + PSPNet head \cite{zhao2017pyramid} & 46.04 & 59.78 & 66.71 & 63.05  \\
        \bottomrule
        - Perfect Filled Images & 43.83 & 64.92 & 57.43 & 60.94  \\
        \midrule
        - Pretrained Weights & 44.93 & 66.22 & 58.29 & 62.00   \\
        \midrule
        + Hole Mask & 45.96 & 66.07 & 60.16 & 62.98 \\
        \midrule
        + Pseudo Pretraining & 46.44 & 62.01 & 64.91 & 63.43  \\
        \midrule
        \: + Pseudo Pretraining \& Real Images \: & \textbf{46.77} & 59.59 & \textbf{68.49} & \textbf{63.73}  \\
        \bottomrule 
        \: Human Subject A \: & 45.60 & \textbf{75.07} & 53.73 & 62.64 \\
        \midrule
        \: Human Subject B \: & 42.21 & 60.40 & 58.36 & 59.36 \\
        \midrule
        \: Human Subject C \: & 36.85 & 61.47 & 47.93 & 53.86 \\
        \bottomrule 
      \end{tabular}
      }
      \vspace{3 pt}
      \caption{An ablation study of the segmentation model, and human performance.}
      \label{tab:segmentation}
    \end{center}
    \vspace{-30 pt}
\end{table*}

In the ablation study, we first tried out a few advance segmentation backbones/heads, such as HRNet \cite{wang2020deep} head, PSPNet \cite{zhao2017pyramid} head, ResNet-50 backbone \cite{he2016deep}, and Swin Transformer \cite{liu2021swin} backbone, as shown in the top 3 rows (excluding the header) of Table \ref{tab:segmentation}. However, we do not observe obvious improvement when using the more complex backbones or heads for our task, after several trials of training comparison. We think a major potential reason is that our segmentation performance of the simpler backbone \cite{he2016deep} and head \cite{zhao2017pyramid} is nearly saturated given the highly subjective labels, and thus simply adding capacity or complexity of backbone does not improves much. This is discussed more in details when we compare with human performance in section 4.2. Thus, we chose ResNet-50 backbone \cite{he2016deep} + PSPNet head \cite{zhao2017pyramid} as our base network, due to its simplicity and efficiency. The rest of ablation studies all shared the same base network for fair comparison, and thus the results should be compared with $3^{rd}$ row. 


Besides the network backbones, we also studied other aspects that might potentially affect the segmentation performance. As we mentioned in section \ref{data_labeling}, 832 images in the labeled dataset have almost perfect fill and thus have no mask labeling, and thus we wonder whether having these images in the training would be helpful. In the $4^{th}$ row, we can see that the model trained without using these images indeed has worse performance, which concludes that adding perfect fills to training is important. All of our models starts training based on the checkpoints pretrained on ADE20K \cite{zhou2017scene}, and we show that the performance also decreases obviously without pretrained weights, as shown in the $5^{th}$ row. Another intuitive thing is to concatenate the hole mask in the input, as it could theoretically help the network quickly localize the potential artifacts region. However, as shown in the $6^{th}$ row, our experiments show that adding the mask into input channel does not actually boost the segmentation performance, and thus we decide not to use it for the simplicity purpose.

We also studied the possibility of generating pseudo labels on large scale unlabeled images for the pretraining purpose. Inspired by BoxInst \cite{tian2021boxinst}, which used bounding box masks as weak supervision to train instance segmentation, we aim to find some similar "enlarged" masks covering the artifacts region as our pseudo labels. Initially, we tried using the hole mask as weak supervision, but realize that the network quickly overfits on the high-frequency artifacts on the hole boundary, which is not useful for our purpose. To this end, we used a pretrained artifacts segmentation network to generate artifacts mask regions on 100K unlabeled images, and then enlarged the segmented masks by some random dilation iterations to cover the perceptual artifacts region. The results in the $7^{th}$ row show that such pretraining strategy slight improves the performance. Finally, we also tried adding the same quantity of real images into training, where the masks are empty for these real images. The $8^{th}$ row shows that this is also useful to further boost the performance. 

\begin{figure*}[!h]
    \centering
    \vspace{-15 pt}
    \includegraphics[trim=0.5in 1.0in 0.5in 0in, clip,width=\textwidth]{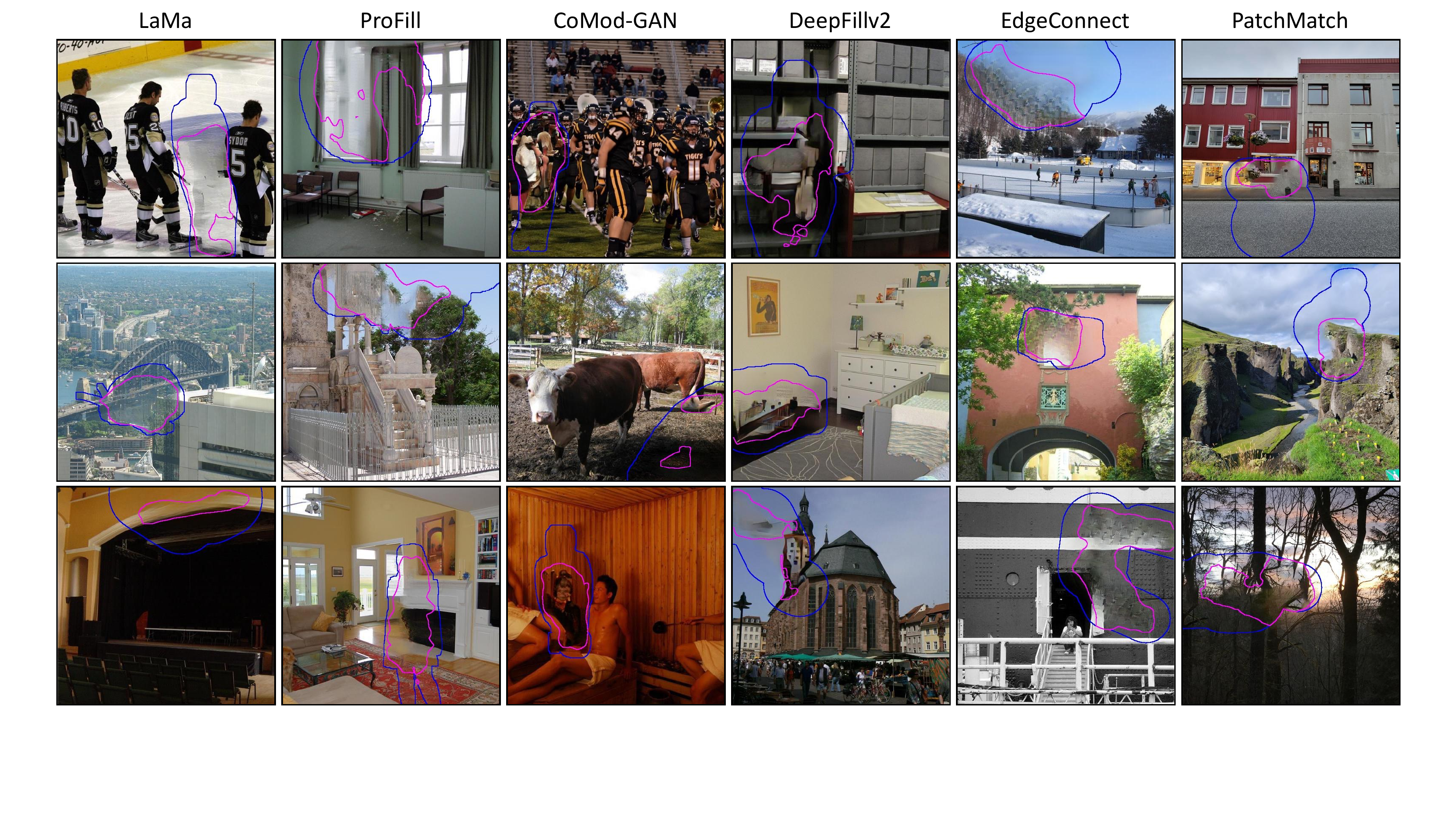}
    \vspace{-20 pt}
    \caption{Qualitative results of the predicted bad fill segmentation on six different inpainting model outputs. The pink and blue boundaries indicate the predicted bad fill region and the hole region, respectively. Please feel free to zoom in to see the details.  }
    \label{fig:seg_results}
    \vspace{-30 pt}
\end{figure*}


\vspace{-5pt}
\subsection{Analysis on Human Perceptual Judgement} 
\vspace{-5pt}

As mentioned in section 3, the labeled dataset from our labeling team has been carefully checked and thus have relatively high quality. In order to better understand how subjective human judgements are and the human performance bound, we ask three more human subjects to label on the 479 test images. Then, we compare the labels from these three human subjects with the previous labels from the labeling team, which are shown in the last three rows of Table \ref{tab:segmentation}. Regarding the three workers' background, human subject A has worked on this task before but not these images, and human subjects B \& C have never worked on this task before but are taught by the labeling team with a bunch of labeled examples. Thus, human subject A should theoretically have better understanding of the task as well as the labeling criterion of the labeling team, compared to the other two subjects. All of them have photography or design background. 

Interestingly, the results show that our segmentation model reaches and even surpasses the best human subject on all metrics except for precision. This infers that our model actually learns a better understanding of averaged judgement criterion of the labeling team, compared to each individual human. On the other hand, these results also indicate that humans have very subjective opinions on the labeling the artifacts regions, as the quantitative scores deviate obviously from each other. A visual illustration of different people's labels on the same filled image is shown in Fig. \ref{fig:human_label_comparison}, and we include more examples like this in the supplemental. Since our segmentation performance surpasses the human performance, this indicates that our segmentation model reaches to a near saturation point for this highly subjective segmentation task. This might also explain why more complex backbone \cite{liu2021swin} or other tweaks of data or training do not provide significant performance improvement, as we observed in the ablation study.

\begin{figure*}[h]
    \centering
    \vspace{-15 pt}
    \includegraphics[trim=0.0in 4.8in 0.0in 0in, clip,width=\textwidth]{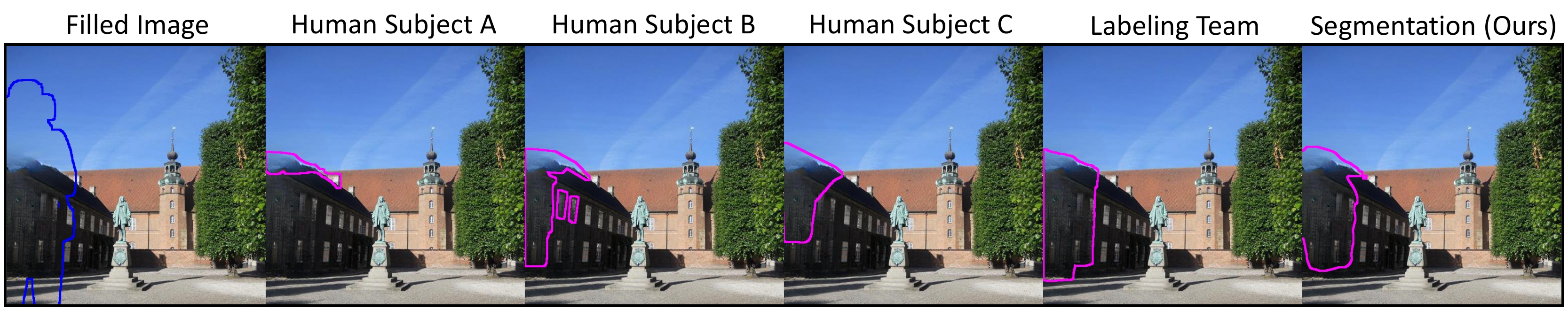}
    \vspace{-20 pt}
    \caption{A visual comparison between labels from multiple human subjects on the same filled image. Our segmentation result is shown in the last column. }
    \label{fig:human_label_comparison}
    \vspace{-25 pt}
\end{figure*}


\vspace{-5pt}
\section{Evaluating Inpainting Quality for Object Removal}
\vspace{-5pt}




\subsection{Motivation}

It has been widely discussed \cite{yu2018generative}\cite{yu2019free} that image inpainting lacks good evaluation metrics, especially for single image quality assessment. Previous works mostly treat image inpainting as a kind of restoration task, so the reconstruction metrics, such as MSE, SSIM, PSNR, and LPIPS \cite{zhang2018unreasonable}, are often used to quantify the similarity between the filled image and the original image. Thinking carefully, we realize that reconstruction metrics might reasonably measure inpainting performance only when the holes are not very large and on the background region. When the holes largely overlap with or cover the foreground objects, most inpainitng algorithms would fill the hole regions by using the background context, where the object is oftentimes completely removed from the image. In these scenarios, reconstruction metrics are no longer proper metrics to gauge the inpainting quality, since the filled region could be totally irrelevant to the original pixels inside the hole. As shown in Fig. \ref{tab:eval_metrics}, when removing the person from the image, output A is visually more plausible than output B, but somehow all the existing reconstruction metrics make opposite judgement. Embarrassingly, object removal is arguably the most frequently used applicable scenario for inpainting algorithms. Thus, it means we really lack good metric for assessing inpainting quality in this scenario. This motivates us to think if the perceptual artifacts localization could be used as a no-reference metric to evaluate inpainting quality in the object removal scenario. 

\begin{figure*}[h]
    \vspace{-10 pt}
    \centering
    \includegraphics[trim=0.2in 0.8in 0.2in 0in, clip,width=\textwidth]{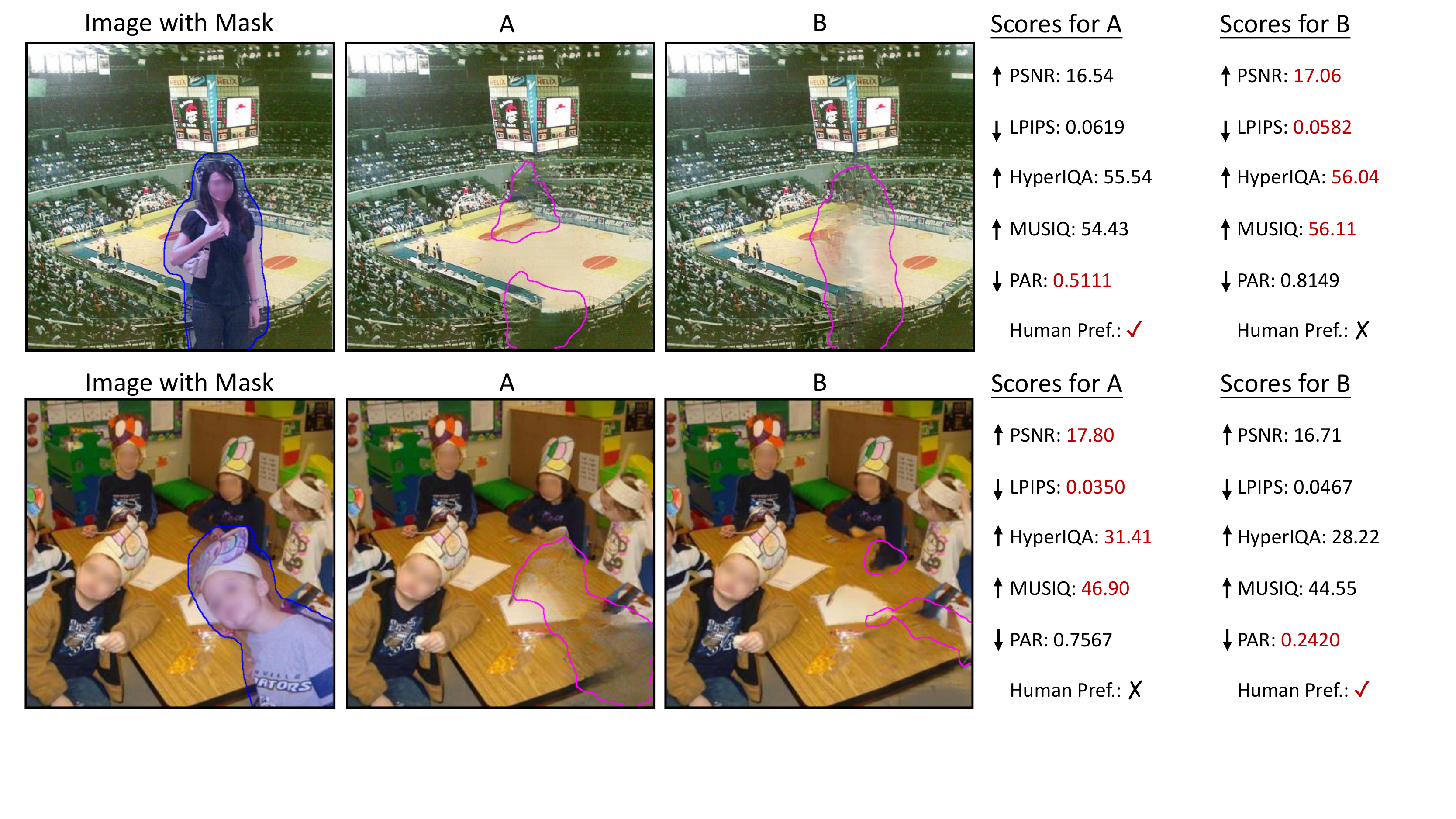}
    \vspace{-25 pt}
    \caption{A visual illustration of filled outputs by two inpainting models \cite{zeng2020high}\cite{nazeri2019edgeconnect}, with the corresponding metric scores. The red scores indicate the preferred choice according to each metric. }
    \label{fig:data_labeling}
    \vspace{-15 pt}
\end{figure*}

\subsection{Metric Definition}

Since our segmentation model could generalize reasonably well to diverse and unseen inpainting methods, we start to wonder whether the size of the detected artifacts region can be used as a metric to assess the inpainting quality. Basically, we assume that an image with good inpainting quality should have relatively smaller the perceptual artifacts region, and vice versa. We name this metric as Perceptual Artifacts Ratio (PAR), which is the ratio of "size of the perceptual artifact region / size of the input hole". The metric computation procedure is that we first run the segmentation model on the filled image and then compute PAR for any filled images, without the need of using the original image. During the quality comparison between different inpainting models, we simply evaluate which inpainting ouptuts have smaller artifacts region among the comparisons.


\subsection{Correlation with Human Perception}

In order to evaluate how well our PAR metric correlates with human perception, we collected user preferences on the filled images between four pairs of inpainting methods. Among these user comparisons, two pairs of comparisons happen between two strong inpainting models, as shown in the first two rows of Table. \ref{tab:eval_metrics}, and another two pairs are between one strong and one relatively weak model, as shown in the bottom two rows of Table. \ref{tab:eval_metrics}. In each pair of methods comparison, we show users two filled images with randomized order, and ask users to pick the preferred image out of the two options. The user studies were conducted on Amazon Mechanical Turk (AMT), where we asked five users to vote on each image. Finally, we consider that one filled image is strongly preferred than the other, only if 4 out of 5 users reach an agreement. In this study, we only used the strongly preferred image pairs as human preference ground truth to reduce the noise as much as possible, where the number of strongly preferred cases are shown in the $2^{nd}$ column of Table \ref{tab:eval_metrics}. Since we are evaluating inpainting quality in the object removal scenarios, we use Mask R-CNN \cite{he2017mask} pretrained on COCO \cite{lin2014microsoft} to generate object masks, and dilate three iterations with $5\times5$ kernel to increase the mask coverage on the object. 

As shown in Table \ref{tab:eval_metrics}, out of 1,000 images for each pair of method comparison, we found that users reach strong agreement on a subset of images with quantity ranging from 321 to 718 shown in the $2^{nd}$ column. The reason why the number of strongly preferred cases of "LaMa vs. ProFill" are less than the others is that these two methods have relatively closer inpainting performance, which causes more disagreement. Other columns in Table \ref{tab:eval_metrics} basically indicates the percentage of correct ranking from each metric, with respect to the human perceptual judgement. In this study, we compare with two reconstruction metrics PSNR and LPIPS \cite{zhang2018unreasonable}, as well as  two NR-IQA metrics Hyper-IQA \cite{su2020blindly} and MUSIQ \cite{ke2021musiq}. Overall, the quantitative results show that our PAR metric outperforms all these existing metrics for assessing inpainting quality in object removal scenarios. 

\begin{table*}[t!]
    \begin{center}
     \resizebox{\textwidth}{!}{
      \begin{tabular}
      {c|c|c|c|c|c|c}
        \toprule 
        \textbf{Comparisons} 
                          & \ \hspace{2pt} No. Pairs \hspace{2pt}
                          & \ \hspace{2pt} PSNR \hspace{2pt}
                          & \ \hspace{2pt} LPIPS \cite{zhang2018unreasonable} \hspace{2pt}
                          & \ \hspace{2pt} HyperIQA \cite{su2020blindly} \hspace{2pt}
                          & \ \hspace{2pt} MUSIQ \cite{ke2021musiq} \hspace{2pt}
                          & \ \hspace{0pt} PAR (Ours) \hspace{0pt} \\
        \midrule 
        LaMa vs. ProFill & 321 & 56.70 \% & 62.31 \% & 39.97\% & 65.11\% & \textbf{65.42 \%}  \\
        \midrule
        LaMa vs. CoMod-GAN & 367 & 48.77 \% & 48.77 \% & 51.50\% & 55.31\% & \textbf{69.21 \%}   \\
        \midrule
        ProFill vs. EdgeConnect & 560 & 23.92 \% & 11.96 \% & 56.39\% & 49.62\% & \textbf{79.82 \%}  \\
        \midrule
        LaMa vs. EdgeConnect & 718 & 44.71 \% & 43.45 \% & 35.71\% & 71.72\% & \textbf{72.70 \%}  \\
        \bottomrule
        Overall & 1966 & 41.50\% & 38.55 \% & 45.24\% & 61.28\% & \textbf{72.89 \%}  \\
        \bottomrule 
      \end{tabular}
      }
      \vspace{5 pt}
      \caption{Quantitative results for measuring the correlation between different metrics and human perceptual judgement. }
      \label{tab:eval_metrics}
    \end{center}
    \vspace{-35 pt}
\end{table*}

\begin{figure*}[!h]
    \centering
    \vspace{-15 pt}
    \includegraphics[trim=0.0in 3.6in 1.1in 0in, clip,width=\textwidth]{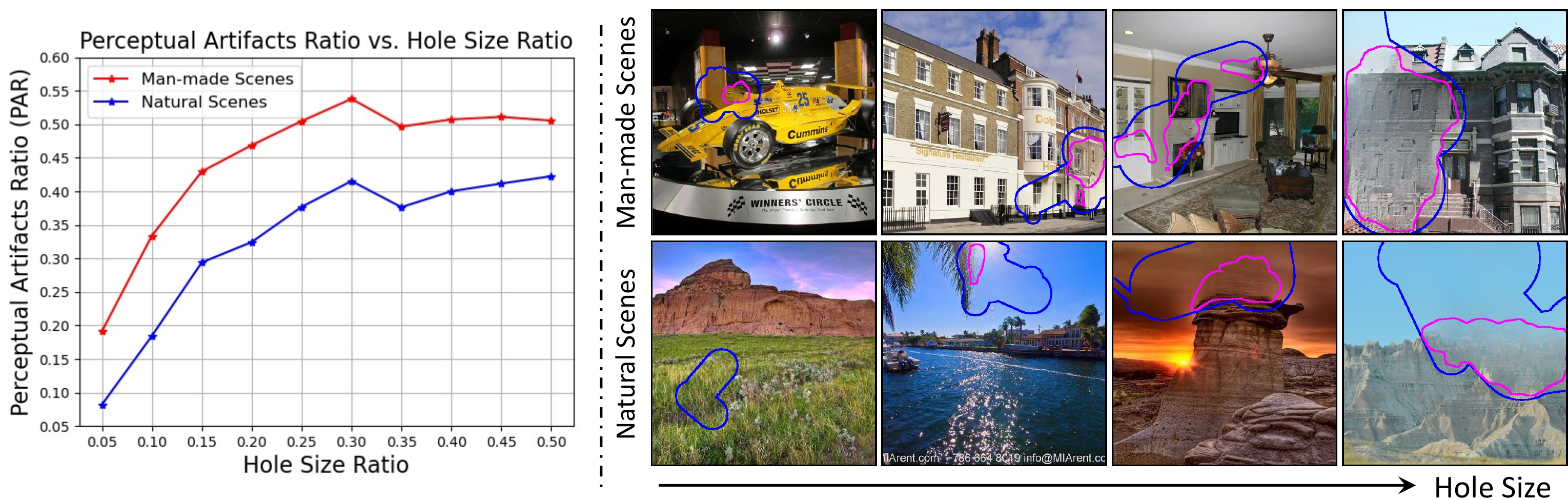}
    \vspace{-20 pt}
    \caption{\textbf{Left}: relationship between PAR and hole size for both man-made scenes and natural scenes. \textbf{Right}: some visual examples of segmented perceptual artifacts region (pink boundary) with varying hole (blue boundary) size. Inpainting models like LaMa produce more artifacts when the hole is larger and scenes are more complex. }
    \label{fig:PAR_vs_HoleSize}
    \vspace{-25 pt}
\end{figure*}

\subsection{PAR Analysis with Hole Size and Scene Types}




We claimed that inpainting artifacts mostly appear in larger holes and complex scene structures. Using our pretrained artifacts segmentation model, we also studied how PAR would change with respect to the hole size in two scenarios: man-made scenes and natural scenes. In the places2 testing dataset, we sampled man-made scenes from the categories, such as building, room, shop, stadium, studio, factory and so on. On the other hand, natural scenes are sampled from categories, such as sky, land, mountain, forest, garden, pasture, beach, desert, and so on. We sampled 2,000 test images for both scenarios and randomly placed stroke holes of specific sizes on them. Then we run LaMa to fill the hole. The relationship between PAR and hole size for natural or man-made scenes is shown in Fig. \ref{fig:PAR_vs_HoleSize}. Our conclusions from the figure are: (1) As the hole size increases, LaMa has a higher possibility of generating inpainting artifacts. (2) Inpainting models like Lama struggles more to complete man-made structures than natural scenes. We believed that this rule applies to other inpainting algorithms as well.

\vspace{-10pt}
\section{Making Inpainting Models Iterative}
\vspace{-5pt}





Modern inpainting algorithms have shown consistent performance improvement over the last few years. However, when inpainting large holes, we often still observe that the inpainting models could often perfectly fill a partial region of the hole while generating obvious artifacts on the other regions. Given this observation, an intuitive idea is that: if the perceptual artifacts region can be reliably segmented out, can we enable the inpainting refill on the artifacts region? In this section, we discuss how we make the inpainting models iteratively fill on the artifacts region, and its effectiveness to improve the inpainting quality.

\vspace{-5pt}
\subsection{Iterative Fill Pipeline}
\vspace{-2pt}

\begin{figure*}[!h]
    \centering
    \vspace{-20 pt}
    \includegraphics[trim=0in 4.3in 0.0in 0in, clip,width=\textwidth]{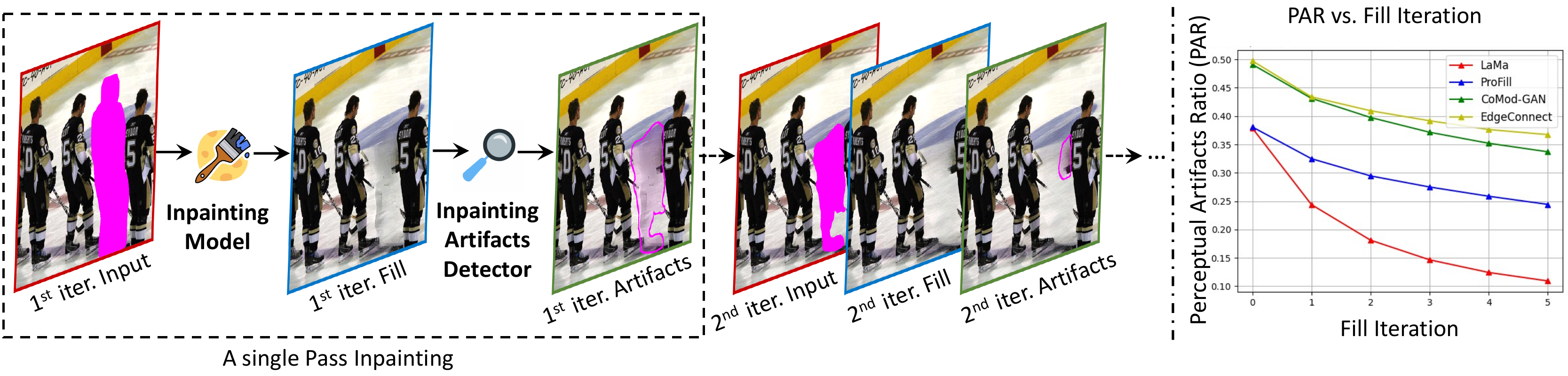}
    \vspace{-20 pt}
    \caption{\textbf{Left}: an overview pipeline of our iterative fill. \textbf{Right}: curves that show predicted perceptual artifacts ratio consistently decreases over the fill iteration for all inpainting models. }
    \label{fig:iter_fill_pipeline}
    \vspace{-25 pt}
\end{figure*}

In Fig. \ref{fig:iter_fill_pipeline}, we show an overview of our iterative fill pipeline. The input image with hole is first fed into an inpainting model to generate a filled image. Then, the filled image is fed into our perceptual artifacts segmentation model to detect the artifacts region, which are converted into the hole mask for the next iteration inpainting. We post-process the segmentation output of artifacts region by multiplying it with the original hole mask in an element-wise manner, so that we ensure not to change any pixels outside the original hole during iterative fill. Our iterative fill pipeline is extremely simple to integrate with and agnostic to all the inpainting models. 

\begin{figure*}[!h]
    \centering
    \includegraphics[trim=0.4in 0.0in 0.0in 0in, clip,width=\textwidth]{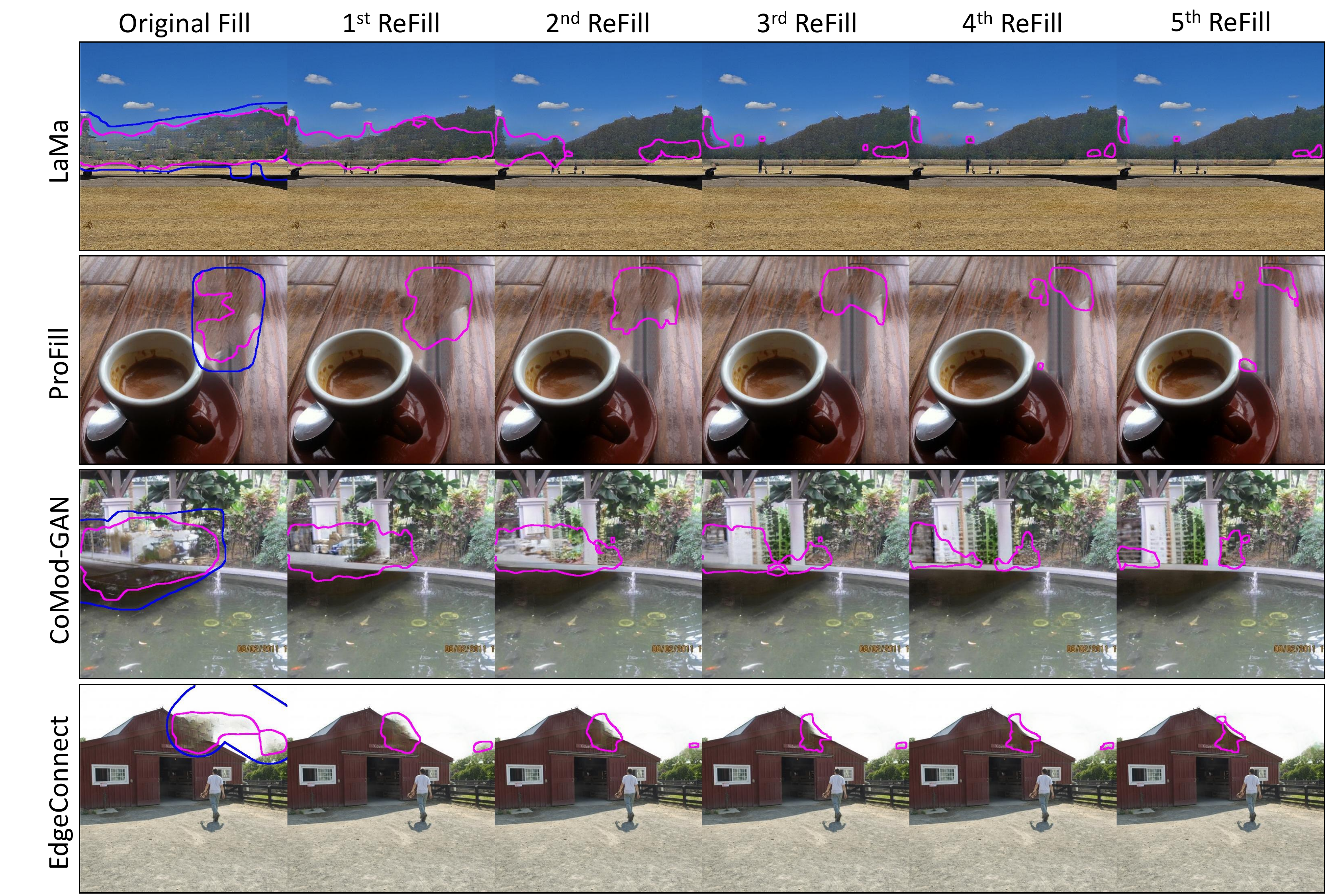}
    \vspace{-20 pt}
    \caption{Qualitative results for iteratively fill of four deep inpainting models. The pink and blue boundary indicate the predicted bad fill region and the hole mask, respectively.  }
    \label{fig:iterative_fill}
    \vspace{-15 pt}
\end{figure*}

\vspace{-5 pt}
\subsection{Performance Improvement by Iterative Fill}
\vspace{-5pt}

We evaluate the performance of the iterative fill from two aspects. First, we compute the size of the detected artifacts region or PAR over the fill iterations, as shown in the right of Fig. \ref{fig:iter_fill_pipeline}. We observe that the detected artifacts region consistently decreases as more iterative fill happens. This indicates that our iterative fill indeed improves the filled image quality, such that less perceptual artifacts are detected. Here, we show up to $5^{th}$ iterative fill in the main paper, and put analysis on more iterations of refill in the supplemental. We also conducted a user study that shows whether users think the $5^{th}$ iteratively filled image are better, same, or worse than the original fill for four inpainting models. As shown in Table \ref{tab:user_study}, our iterative fill pipeline improves approximately 30\% of images compared to the original fill, and rarely make the filled images worse, especially for the best model LaMa \cite{su2020blindly}. This implies that our system can be safely integrated into these inpainting models to boost inpainting quality. All of our user studies are conducted on AMT. In each inpainting method, we uniformly sampled 500 images from the testset, which result in 2,000 images in total. We asked 20 turkers to carefully check on each image and averaged the preference. We do not use the traditional metrics to quantify the performance between original fill and iterative fill, since we found that these metric scores between them are too close and sometimes random, which does not reflect much information. We have more discussion on this in the supplemental.

\begin{figure*}[!h]
    \centering
    \vspace{-20 pt}
    \includegraphics[trim=0in 0.0in 2.6in 0in, clip,width=\textwidth]{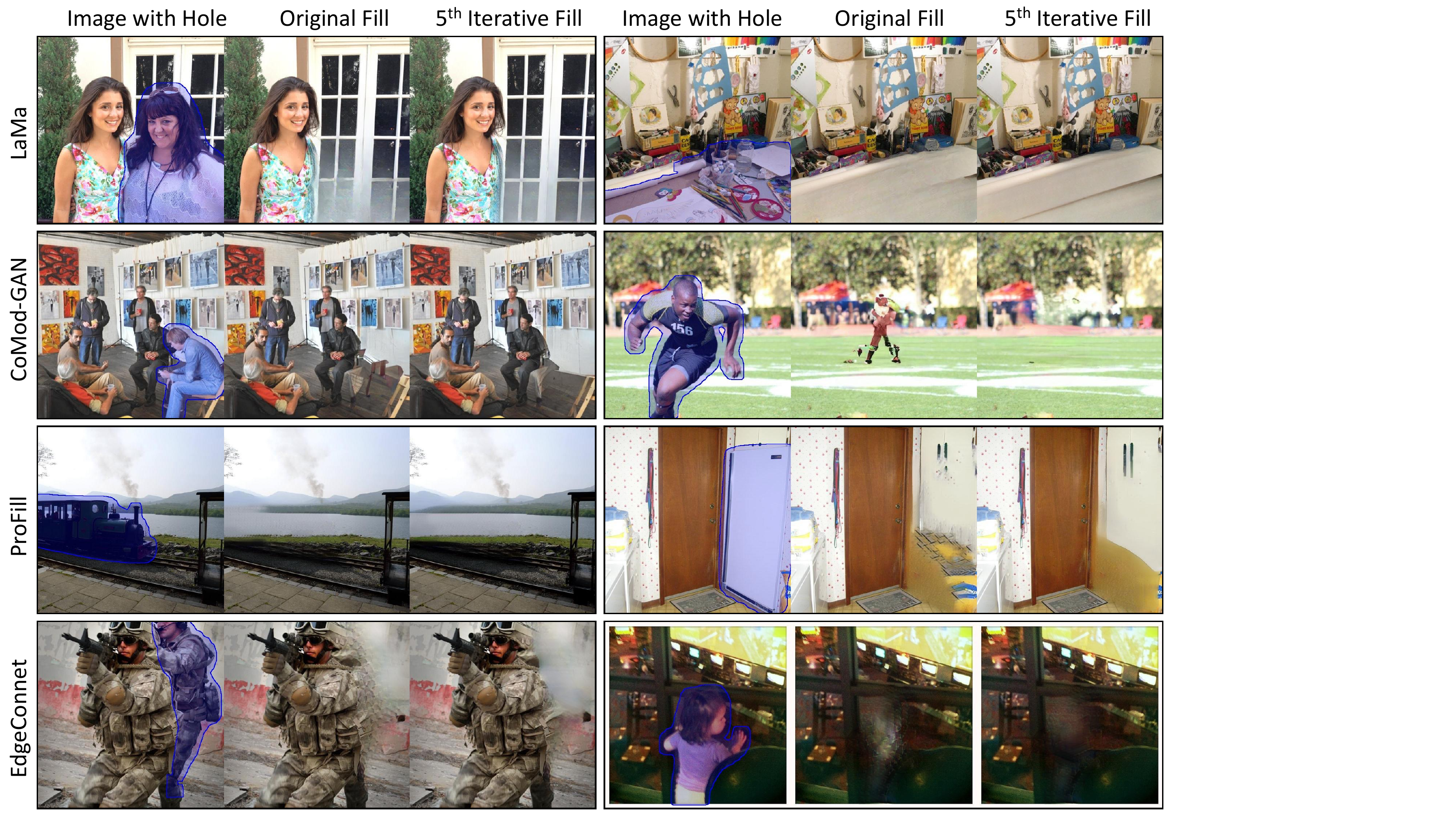}
    \vspace{-20 pt}
    \caption{Qualitative comparison between the original fill and the $5^{th}$ iterative fill. }
    \label{fig:fill_vs_5threfill}
    \vspace{-30 pt}
\end{figure*}

In Fig. \ref{fig:iterative_fill}, we show a few qualitative examples of how filled image improves over a number of iterative fills, and more visual comparison between original fill and $5^{th}$ iterative fill in Fig. \ref{fig:fill_vs_5threfill}. We observe that iterative fill could oftentimes help the inpainting models refine both structure and texture in many cases. However, due to the limitation of the inpainting algorithms themselves, the predicted perceptual artifact regions would not always reach to zero and thus would still leave some artifacts in the image. 
Finally, we also compared with a heuristic approach to gradually erode hole mask five times and iteratively fill the hole using the eroded masks in a onion-peel fashion similarly to \cite{zeng2020high,oh2019onion,guo2019progressive}, which we name this heuristic as onion fill. The results, as shown in Table. \ref{tab:onion_fill}, indicate that our iterative fill on the perceptual artifacts region are much more preferred by users, compared to the iterative fill on the heuristically eroded mask. In this study, we use LaMa \cite{su2020blindly} for both types of iterative fill. 

We show visual comparison between original fill and $5^{th}$ iterative fill in Fig. \ref{fig:fill_vs_5threfill}. We observe that iterative fill could oftentimes help the inpainting models refine both structure and texture in many cases. However, due to the limitation of the inpainting algorithms themselves, the predicted perceptual artifact regions would not always reach to zero and thus would still leave some artifacts in the image.

\begin{table*}[h!]
 \vspace{-10pt}
    \begin{center}
     \resizebox{0.9\textwidth}{!}{
      \begin{tabular}
      {c|c|c|c}
        \toprule 
        \textbf{Models} & \ \hspace{10pt} Preferred Original Fill \hspace{10pt} & \ \hspace{10pt} Same \hspace{10pt} & \ \hspace{10pt} Preferred Iterative Fill \hspace{10pt}  \\
        \midrule 
        \hspace{10pt} EdgeConnect \hspace{10pt} & \hspace{10pt} 53 (10.6\%) \hspace{10pt} & \hspace{10pt} 258 (51.6\%) \hspace{10pt} & \hspace{10pt} 189 (37.8\%) \hspace{10pt} \\
        \midrule
        CoMod-GAN & 45 (9.0\%) & 334 (66.8\%) & 121(24.2\%) \\
        \midrule
        ProFill & 14 (2.8\%) & 337 (67.4\%) & 149 (29.8\%) \\
        \midrule
        LaMa & 9 (1.8\%) & 341 (68.2\%) & 150 (30.0\%) \\
        \bottomrule 
      \end{tabular}
      }
      \vspace{5 pt}
      \caption{A user study to show the comparison between original fill and the $5^{th}$ refill. }
      \label{tab:user_study}
    \end{center}
    \vspace{-35 pt}
\end{table*}

\begin{table*}[h!]
    \begin{center}
     \resizebox{\textwidth}{!}{
      \begin{tabular}
      {c|c|c|c}
        \toprule 
        \textbf{Models} & \ \hspace{10pt} Preferred Onion Fill \hspace{10pt} & \ \hspace{10pt} Same \hspace{10pt} & \ \hspace{10pt} Preferred Iterative Fill (Ours) \hspace{10pt}  \\
        \midrule 
        Onion Fill vs. Ours & 15 (3\%) & 354 (70.8\%) & 131 (26.2\%) \\
        \bottomrule 
      \end{tabular}
      }
     \vspace{5 pt}
      \caption{A user study to show the comparison between onion fill and our iterative fill. }
      \label{tab:onion_fill}
    \end{center}
\end{table*}

\section{Conclusion}

In this work, we propose a new learning task of automatically segmenting the perceptual artifacts region in the inpainted images. With our carefully labeled dataset, we train a network to reliably localize and segment the perceptual artifacts, which surpass the performance of three human professionals. Due to the reasonably well generalization across different inpainting models, we propose a new interpretable evaluation metric PAR for inpainting quality assessment in the object removal scenarios, which shows a strong correlation with human preference. Finally, we apply our artifacts segmentation output as new mask to enable inpainting models iteratively fill, and achieve better inpainting quality.

\newpage





\title{Supplemental Materials: \\
Perceptual Artifacts Localization for Inpainting} 


\titlerunning{Perceptual Artifacts Localization for Inpainting}


%
\author{Lingzhi Zhang\inst{1} \and
Yuqian Zhou\inst{2} \and Connelly Barnes\inst{2} \and Sohrab Amirghodsi\inst{2} \and \\ Zhe Lin\inst{2} \and Eli Shechtman\inst{2} \and Jianbo Shi\inst{1}}

\authorrunning{Zhang et al.}
%
\institute{University of Pennsylvania \and
Adobe Research \\
}
\maketitle

\vspace{-20 pt}
\section{Mask Generation}
\vspace{-5 pt}


In this work, all of our images are uniformly sampled from Places2 dataset \cite{zhou2017places}. We generate two types of masks for our experiments, which are masks on the background region and masks covering a complete object. We discuss the details of how to generate these masks in below. 

\begin{figure*}[!h]
    \centering
    \vspace{-20 pt}
    \includegraphics[trim=0.0in 4.8in 0.0in 0.0in, clip,width=\textwidth]{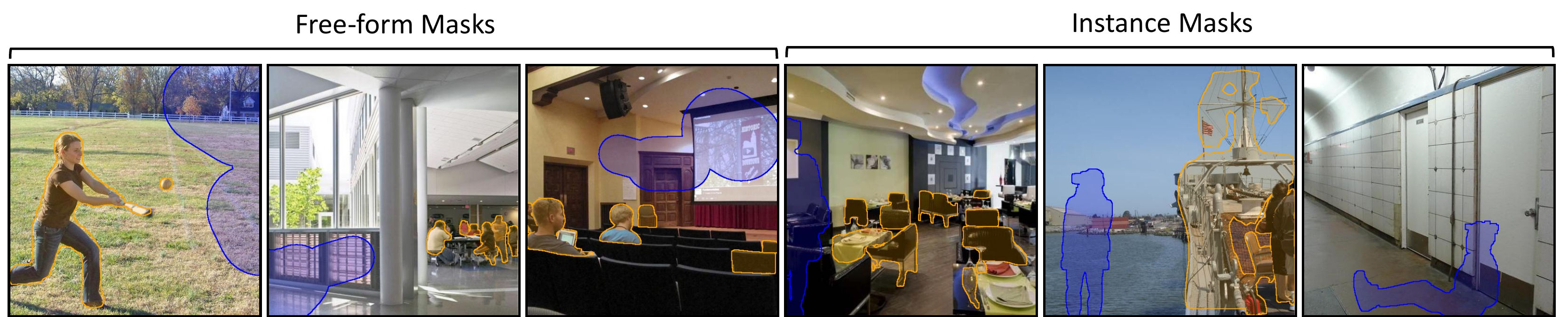}
    \vspace{-20 pt}
    \caption{Some visual examples of mask sampling for labeling and evaluation. The orange masks indicate the foreground region, where the hole masks avoid to overlap with. }
    \label{fig:masks_for_label_and_eval}
    \vspace{-15 pt}
\end{figure*}

\textbf{Masks on the Background.} Since current inpainting models still can not understand the object-level prior and thus can not properly fill the object region, we do not want to sample masks that cover partial objects, which none of the current methods can properly deal with. To this end, we first use Mask R-CNN \cite{he2017mask} to find all object masks, and then avoid sampled holes to partially overlap with these object regions. We use both free-form masks \cite{yu2019free} and instance masks in our experiments, where the hole size ratio over the entire image ranges from 0.08 to 0.3. The instance masks are collected from multiple segmentation datasets, such as COCO \cite{lin2014microsoft} and Pascal VOC \cite{everingham2010pascal}. Some examples of these masks are shown in Fig. \ref{fig:masks_for_label_and_eval}.

\begin{figure*}[!h]
    \centering
    \includegraphics[trim=0.0in 5.3in 0.1in 0.0in, clip,width=\textwidth]{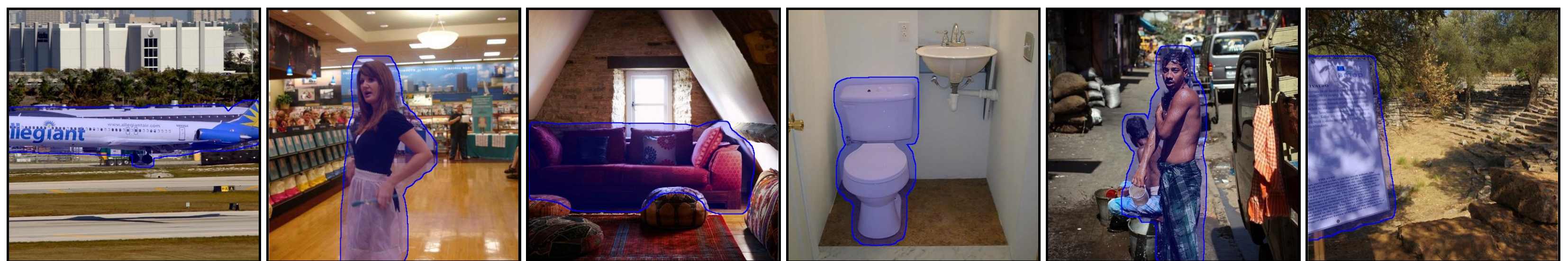}
    \vspace{-20 pt}
    \caption{Some visual examples of mask sampling for object removal inpainting scenarios. }
    \label{fig:masks_for_object_removal}
    \vspace{-10 pt}
\end{figure*}

\textbf{Masks for Object Removal. } In this work, we propose Perceptual Artifacts Ratio (PAR) metric to evaluate the inpainting quality for object removal scenario, which is often consider to be a hard but commonly used scenario for inpainting. To generate the hole mask at scale, we used Mask R-CNN \cite{he2017mask} to first find all object masks in the image, then randomly select one object mask, and dilate the mask with 5 $\times$ 5 kernel for three iterations to increase coverage of the object. Some examples of these object removal masks are shown in Fig. \ref{fig:masks_for_object_removal}.

For our perceptual artifacts labeling, we use a mix of both types of masks. Similarly, in our user study for evaluating "original fill vs. iterative fill" in section 6.2, half of cases use the masks sampled on the background and another half of the cases use the object removal masks. For our PAR metric study in section 5 in the main paper, we use only the masks covering complete objects, since we are evaluating the inpainting quality for object removal scenario.

\vspace{-5 pt}
\section{Training Details of the Segmentation Network}
\vspace{-5 pt}

In this section, we describe the training details of our perceptual artifacts segmentation network. For our final chosen model with ResNet-50 backbone \cite{he2016deep} and PSP head \cite{zhao2017pyramid}, we also added an auxiliary FCN head \cite{long2015fully}, which has a loss ratio of 0.4 compared to the PSP head \cite{zhao2017pyramid}. We train the segmentation network for 20,000 iterations using SGD optimizer with learning rate of 0.01, momentum of 0.9, and weight decay of 0.0005. The learning rate is schedule to decay in polynomial manner by power of 0.9, where the minimum is set to be 0.0001. For the data augmentation, we adopt random flip with a probability of 0.5 as well as JPEG compression. All of our models are trained on 4 NVIDIA RTX-Titan with a batch of 8 on each GPU. All images are trained at 512 $\times$ 512, which is the native resolution of inpainting outputs.

\vspace{-5 pt}
\section{Perceptual Artifacts Labeling Interface}
\vspace{-5 pt}

\begin{figure*}[!h]
    \centering
    \vspace{-10 pt}
    \includegraphics[trim=1.0in 4.0in 1.0in 0.0in, clip,width=\textwidth]{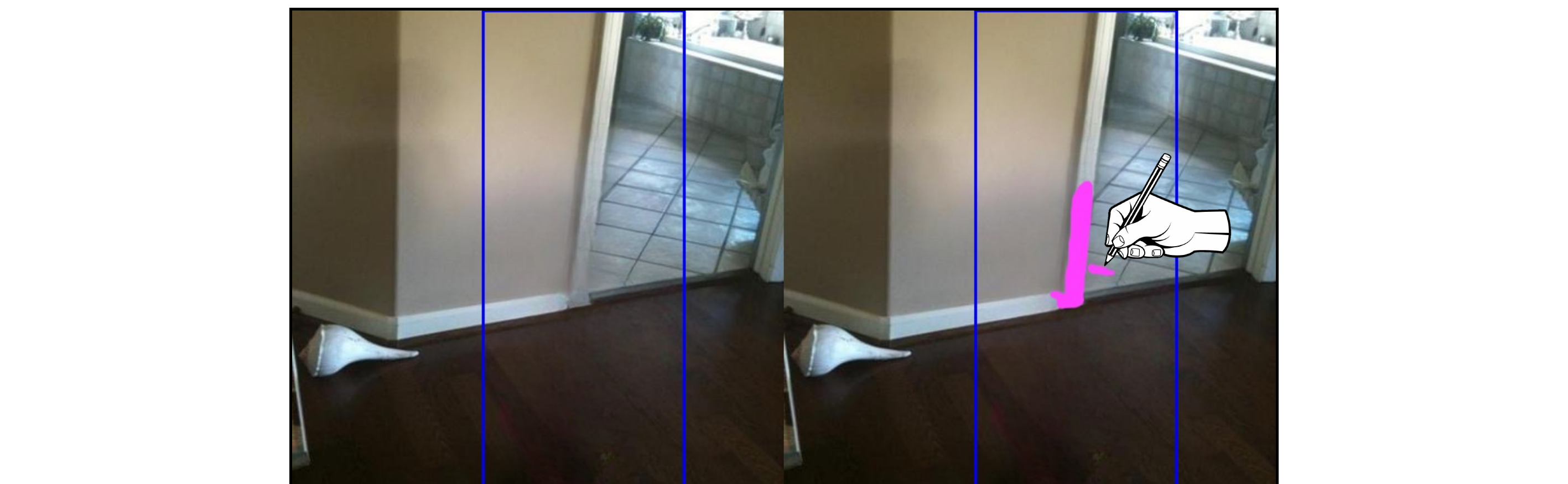}
    \vspace{-20 pt}
    \caption{User labeling interface. }
    \label{fig:user_labeling_interface}
    \vspace{-20 pt}
\end{figure*}

As discussed in the data labeling section in the main paper, we provide a copy of the filled image besides the image that workers actually mark on. The reason is that we want to provide workers a reference to know the original image content, which are useful to judge the perceptual artifacts region. Here, we show an actual worker's interface in Fig. \ref{fig:user_labeling_interface}. 

\vspace{-5 pt}
\section{User Study Interface}
\vspace{-5 pt}

\begin{figure*}[!h]
    \centering
    \vspace{-10 pt}
    \includegraphics[trim=0.0in 3.6in 3.4in 0in, clip,width=\textwidth]{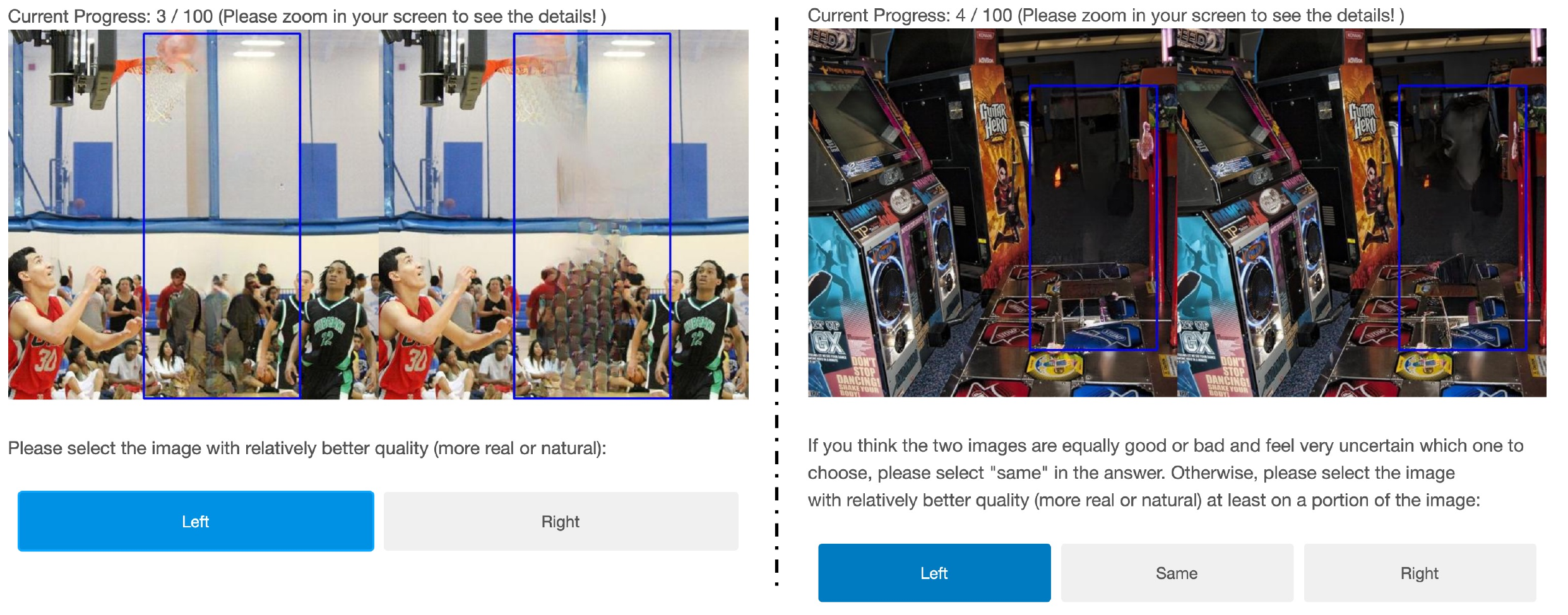}
    \vspace{-20 pt}
    \caption{\textbf{Left}: user study interface for evaluating between two inpainting models, which used for our PAR metric study in section 5.3. \textbf{Right}: user study interface for evaluating whether the iteratively filled images are better than the original fill, which are used as evaluation for section 6.2 in the main paper. Note that left/right images are randomized for each case. }
    \label{fig:user_study_interface}
    \vspace{-30 pt}
\end{figure*}

In this work, we have conducted two user studies using Amazon Mechanical Turk (AMT). In the first user study, we ask users to select the preferred image from two different inpainting methods, where the results are used for evaluating the metric correlation with human in section 5.3 in the main paper. In the second user study, we ask users to choose whether they think the iterative fill is better, same, or worse than the original fill, which are used for evaluation in section 6.2 in the main paper. Here, we show the user interfaces for the first and second studies in the left and right of Fig. \ref{fig:user_study_interface}, respectively. We intentionally do not show the original images to use, since we want to users to judge based on pure perceptual quality without any bias. We use bounding box to indicate the rough hole region instead of the actual hole boundary, so that users would have less bias on the boundary artifacts. 

\section{More Iterations of Fill}

In the main paper, we show how Perceptual Artifacts Ratio (PAR) consistently decreases over the fill iteration up to the $5^{th}$ iteration for 5,000 test images. However, we observe that methods like LaMa \cite{su2020blindly} still have slight tendency of decreasing PAR when increasing the fill iteration. Thus, we run a study by setting the number of iterative fill up to 20 for LaMa \cite{suvorov2021resolution}. As shown in Table. \ref{tab:iterative_fill}, we observe that PAR still consistently decreases over the fill iteration, but the decreasing rate of PAR goes down to a very small number.

\begin{table*}[h!]
  \vspace{-10pt}
    \begin{center}
     \resizebox{\textwidth}{!}{
      \begin{tabular}
      {c|c|c|c|c|c|c|c}
        \toprule 
        \hspace{5pt} \textbf{Iters.} \hspace{5pt} & \ 
        \hspace{5pt} \textbf{PAR} \hspace{5pt} & \ 
        \hspace{5pt} \textbf{Iters.} \hspace{5pt} & \ 
        \hspace{5pt} \textbf{PAR} \hspace{5pt} & \ 
        \hspace{5pt} \textbf{Iters.} \hspace{5pt} & \ 
        \hspace{5pt} \textbf{PAR} \hspace{5pt} & \ 
        \hspace{5pt} \textbf{Iters.} \hspace{5pt} & \ 
        \hspace{5pt} \textbf{PAR} \hspace{5pt} \\
        \midrule 
        1 & 0.3786 & 6 & 0.1091 & 11 & 0.0707 & 16 & 0.0552  \\
        \midrule 
        2 & 0.2439 & 7 & 0.0975 & 12 & 0.0666 & 17 & 0.0533  \\
        \midrule 
        3 & 0.1811 & 8 & 0.0885 & 13 & 0.0628 & 18 & 0.0514   \\
        \midrule 
        4 & 0.1464 & 9 & 0.0814 & 14 & 0.0600 & 19 & 0.0497  \\
        \midrule 
        5 & 0.1241 & 10 & 0.0756 & 15 & 0.0575 & 20 & 0.0481  \\
        \bottomrule 
      \end{tabular}
      }
      \vspace{2 pt}
      \caption{PAR vs. Fill Iters. for LaMa \cite{su2020blindly}.}
      \label{tab:iterative_fill}
    \end{center}
    \vspace{-15 pt}
\end{table*}

\vspace{-5pt}
\section{Why Existing Metrics Are Not Suitable for Comparing ``Original Fill vs. Iterative Fill"?}
\vspace{-10pt}

\begin{figure*}[!h]
    \centering
    \vspace{-20 pt}
    \includegraphics[trim=0.0in 3.9in 0.6in 0.0in, clip,width=0.93\textwidth]{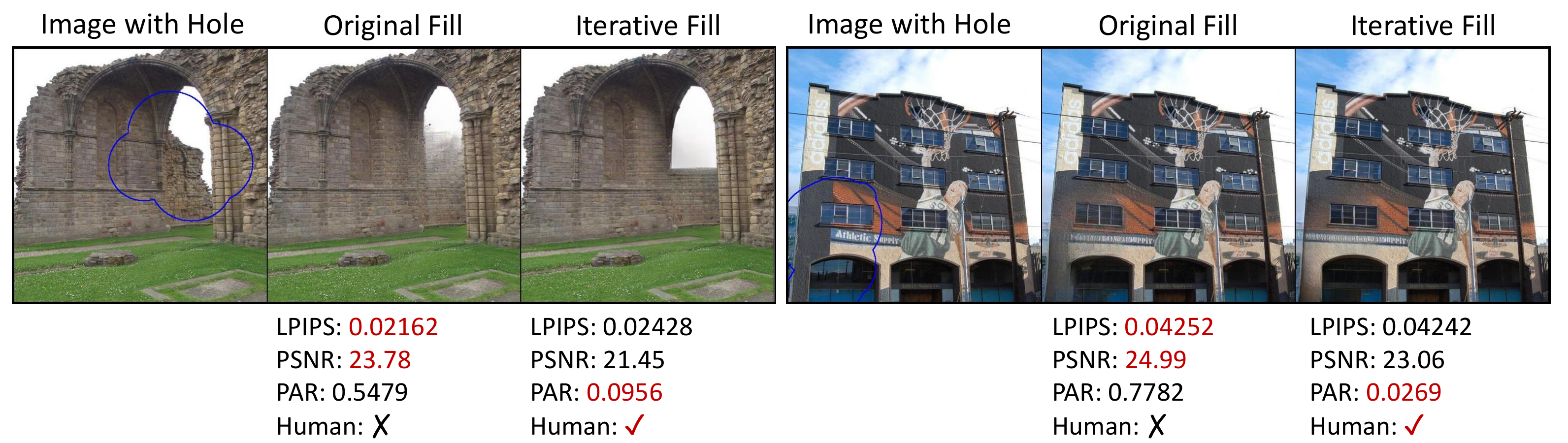}
    \vspace{-10 pt}
    \caption{Metric scores amd human preference for original fill and iterative fill. }
    \label{fig:metrics_not_proper}
    \vspace{-35 pt}
\end{figure*}

While previous works often use metrics, such as LPIPS \cite{zhang2018unreasonable}, PSNR or FID \cite{heusel2017gans}\cite{parmar2021buggy}, to evaluate the performance of different inpainting methods. We found these metrics are not suitable to compare the inpainting qualities between original fill and iterative fill in our case, since the scores are often too similar to make a judgement. The main reason is that since both original fill and iterative fill share the same inpainting algorithm, the difference between their outputs are less obvious than the difference between different inpainting methods, even though the difference might be obvious to human perception. In addition, we found that even when the holes are on the background region, reconstruction metrics LPIPS \cite{zhang2018unreasonable} and PSNR still often prefer the image that is opposite to human judgement, where the typical observations are shown in Fig. \ref{fig:metrics_not_proper}. Thus, we used our proposed PAR metric, which is proven to have strong correlation with human perception in paper's section 4, as well as extensive user studies to evaluate the improvement between original fill and iterative fill, as discussed in section 6.2 in the main paper. 

\section{More Visual Results}
\vspace{-5 pt}

To help readers have better understanding and more insights on our work, we provide more visual examples of human annotations and our predictions of inpainting artifacts, and results of our iterative fill model. 

\vspace{-10pt}
\subsection{Human Labelings}

\begin{figure*}[!h]
    \centering
    \vspace{-10 pt}
    \includegraphics[trim=0.0in 0.2in 0.0in 0in, clip,width=\textwidth]{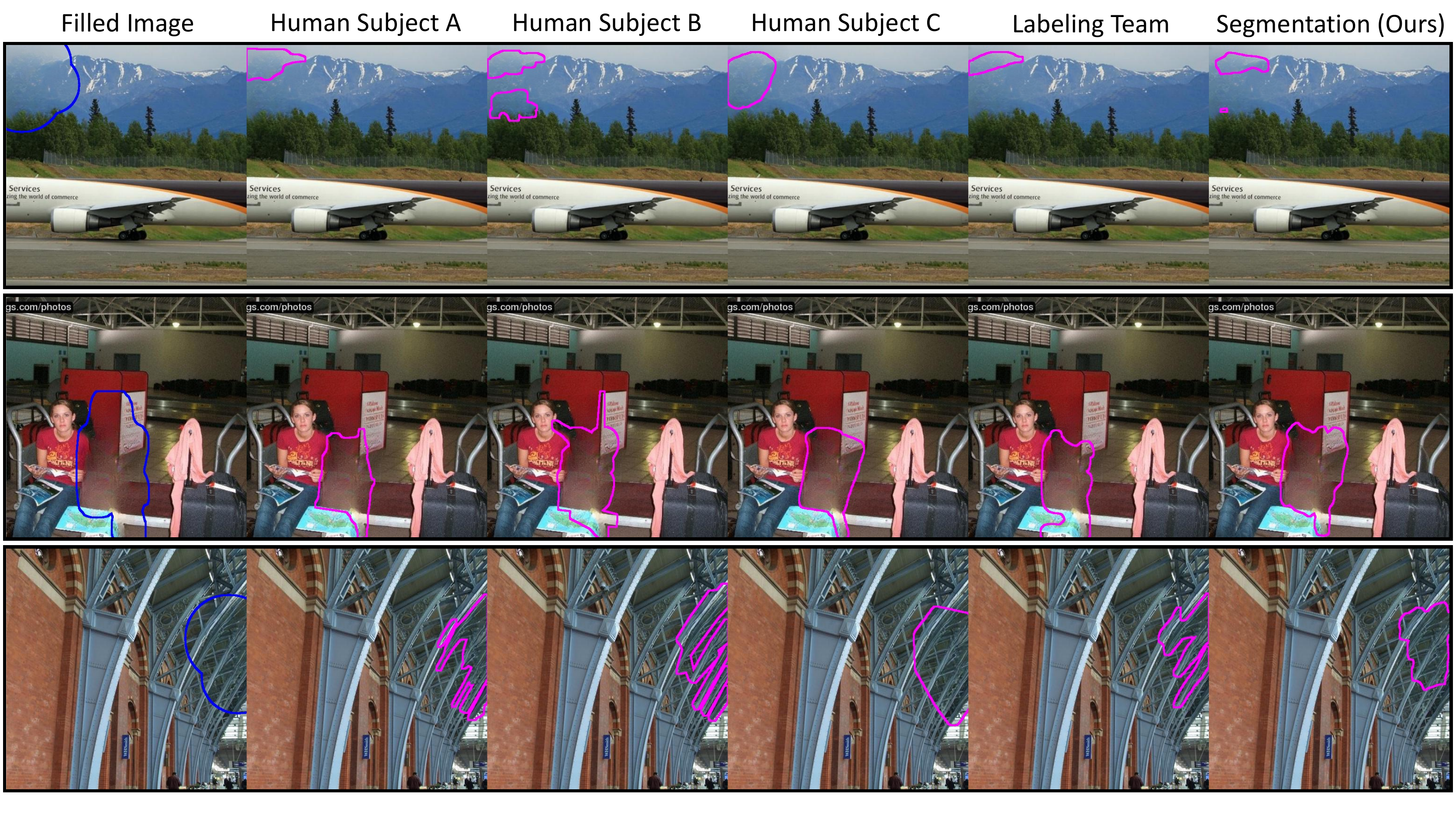}
    \vspace{-20 pt}
    \caption{More qualitative examples of visual comparison between multiple human subjects who label on the same images. The filled images with hole (blue boundary) are shown in the first column, and our segmentation results are shown in the last column. }
    \label{fig:more_human_label_comparison}
    \vspace{-10 pt}
\end{figure*}

\textbf{Visualize Subjective Opinions.} As discussed in section 4.2 in the main paper, labeling perceptual artifacts is a highly subjective task, and different human subjects might have different opinions or standard to label. Here, we show more visual examples of different labels on the same filled images in Fig. \ref{fig:more_human_label_comparison}.

\begin{figure*}[!h]
    \centering
    \vspace{-5 pt}
    \includegraphics[trim=0.0in 0.5in 1.3in 0in, clip,width=\textwidth]{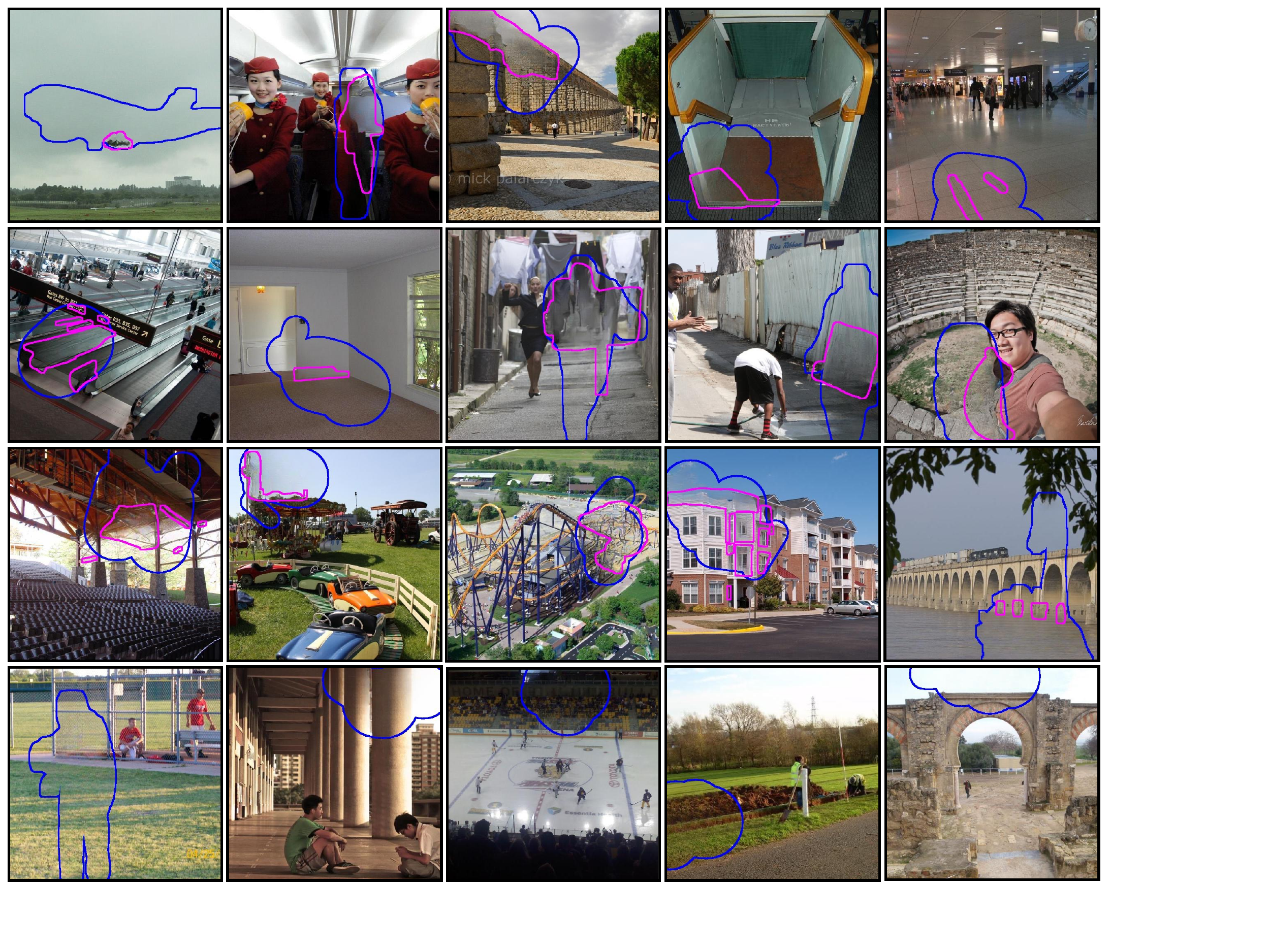}
    \vspace{-20 pt}
    \caption{More qualitative examples of perceptual artifacts labels from the professional human labeling team. The last row shows the perfectly filled images, so the workers do not label anything on them. }
    \label{fig:more_mada_label}
    \vspace{-25 pt}
\end{figure*}

\textbf{More Human Labelings.} In Fig. \ref{fig:more_mada_label}, we show more visual examples of perceptual artifacts labeling from the human professional team. The original hole masks and the artifacts regions are indicated by the blue and pink boundary, respectively. As we mentioned in section 3 in the main paper, there are 832 images that have nearly perfect fills so that human did not put any labels on these images. The visual illustration of these perfect fills are shown in the last row of Fig. \ref{fig:more_mada_label}. We can see that sometimes human's labels go outside of the actual hole mask, since we do not provide hole mask for the workers to avoid any potential bias. However, this won't be an issue, since we intersect the labels with the hole mask to clean up the overly labeled regions as a post-process. 

\newpage

\subsection{Perceptual Artifacts Localization}

\begin{figure*}[!h]
    \centering
    \vspace{-15 pt}
    \includegraphics[trim=0.15in 0.0in 0.5in 0.0in, clip,width=\textwidth]{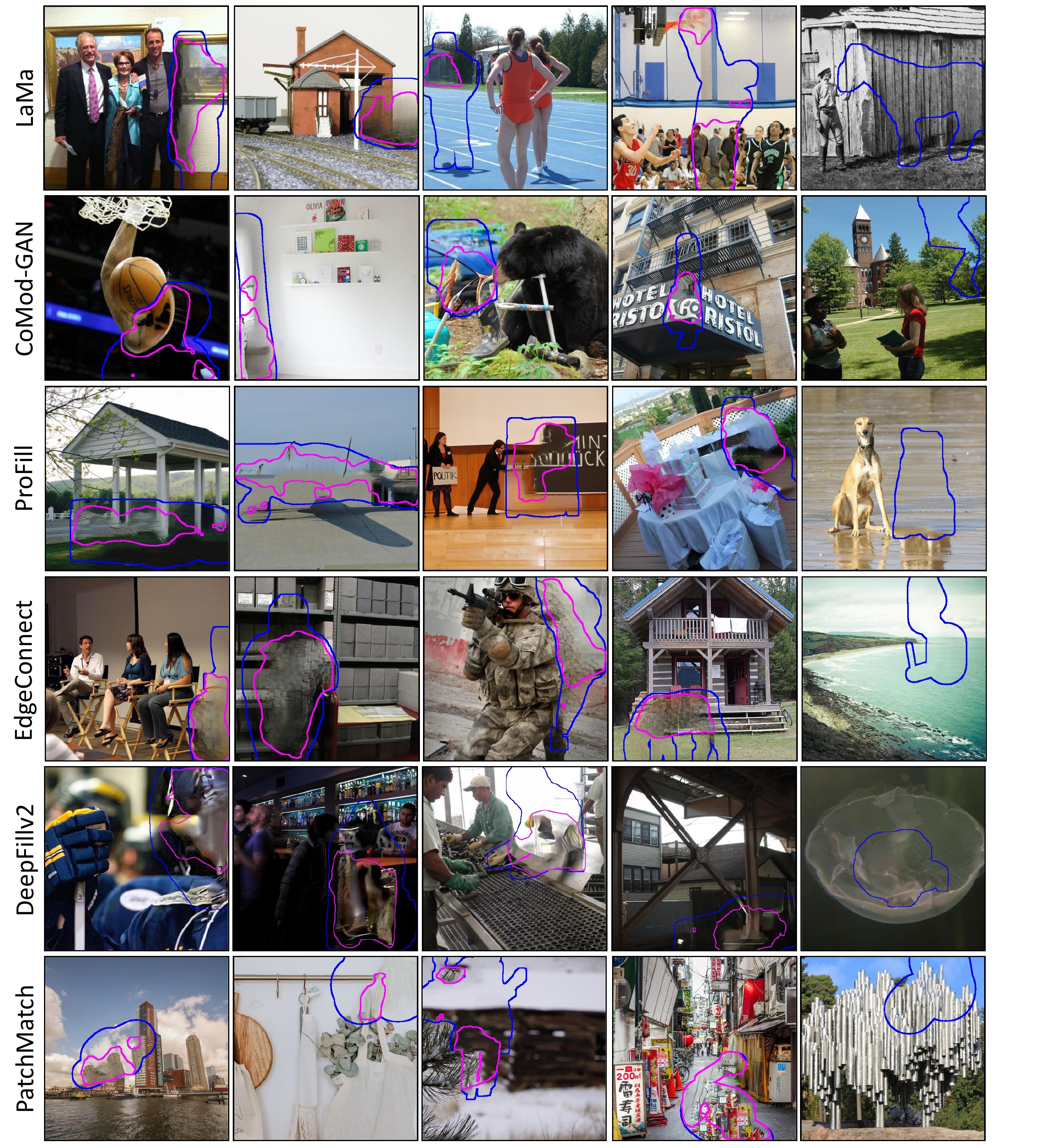}
    \vspace{-20 pt}
    \caption{More qualitative examples of the predicted perceptual artifacts localization from our segmentation network for different inpainting models. In the last column, our network does not predict any mask, since the filled images look almost perfect.}
    \label{fig:more_seg_results}
    \vspace{-10 pt}
\end{figure*}

In Fig. \ref{fig:more_seg_results}, we show more qualitative results of the perceptual artifacts segmentation across different inpainting methods. Note that our models are trained on filled images generated by LaMa \cite{su2020blindly}, CoMod-GAN \cite{zhao2021large}, and ProFill \cite{zeng2020high}. Nevertheless, our trained network generalizes reasonably well to unseen inpainting models, including EdgeConnect \cite{nazeri2019edgeconnect}, DeepFillv2 \cite{yu2019free}, and PatchMatch \cite{barnes2009patchmatch}.

\subsection{Iterative Fill over Iterations}
\vspace{-5pt}

In Fig. \ref{tab:iterative_fill}, we show more examples of iteratively filled images during the process. 

\begin{figure*}[!h]
    \centering
    \vspace{-15 pt}
    \includegraphics[trim=0.1in 7.4in 0.1in 0in, clip,width=\textwidth]{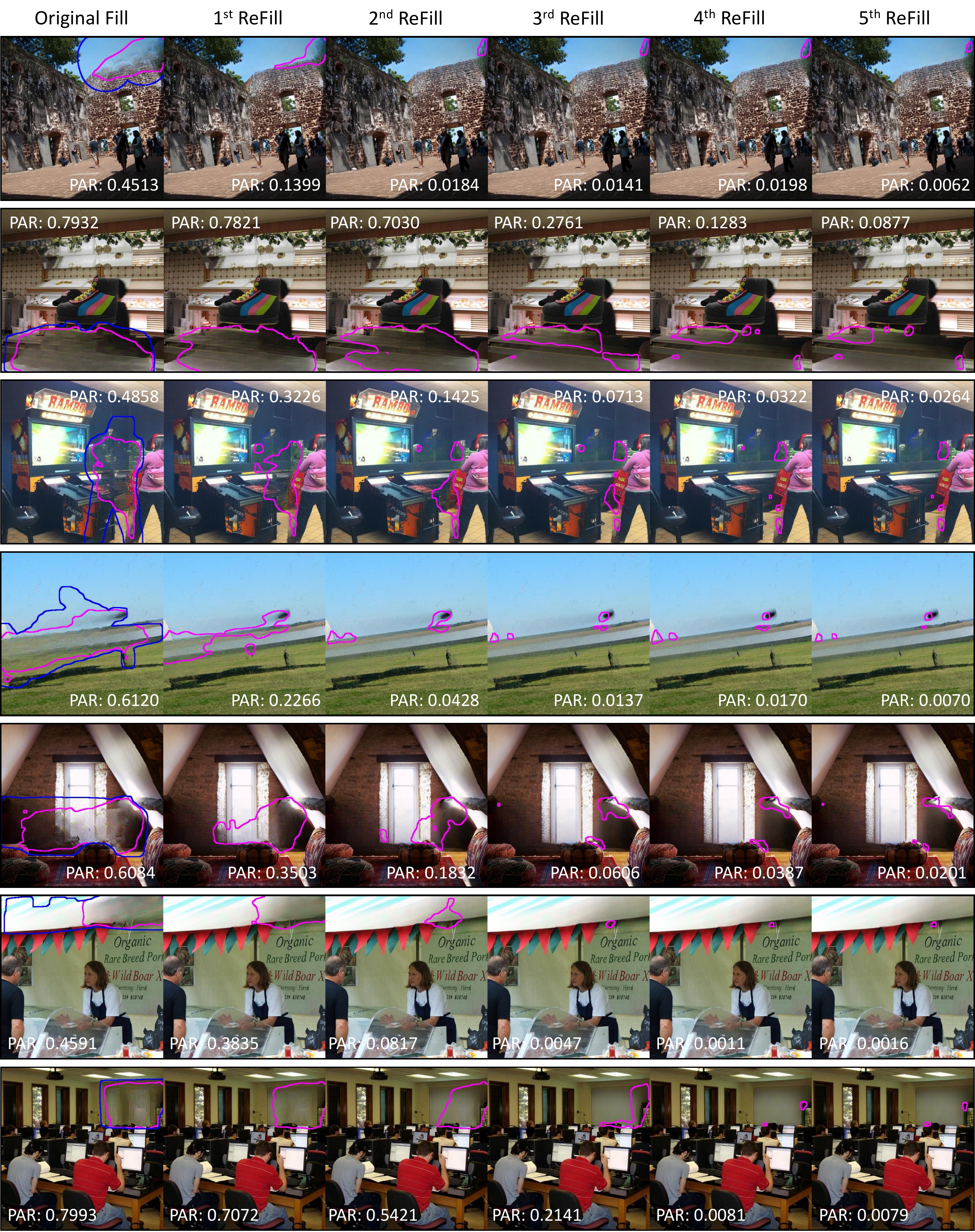}
    \vspace{-20 pt}
    \caption{More qualitative results of iterative fill by LaMa \cite{suvorov2021resolution}. The blue and pink boundaries indicate the original hole mask and perceptual artifacts localization, respetively.  }
    \label{fig:more_iterative_fill}
    \vspace{-20 pt}
\end{figure*}

\vspace{-15pt}
\subsection{Original Fill vs. Iterative Fill}
\vspace{-5pt}

\begin{figure*}[!h]
    \centering
    \vspace{-10 pt}
    \includegraphics[trim=0.0in 0.3in 3.2in 0.0in, clip,width=\textwidth]{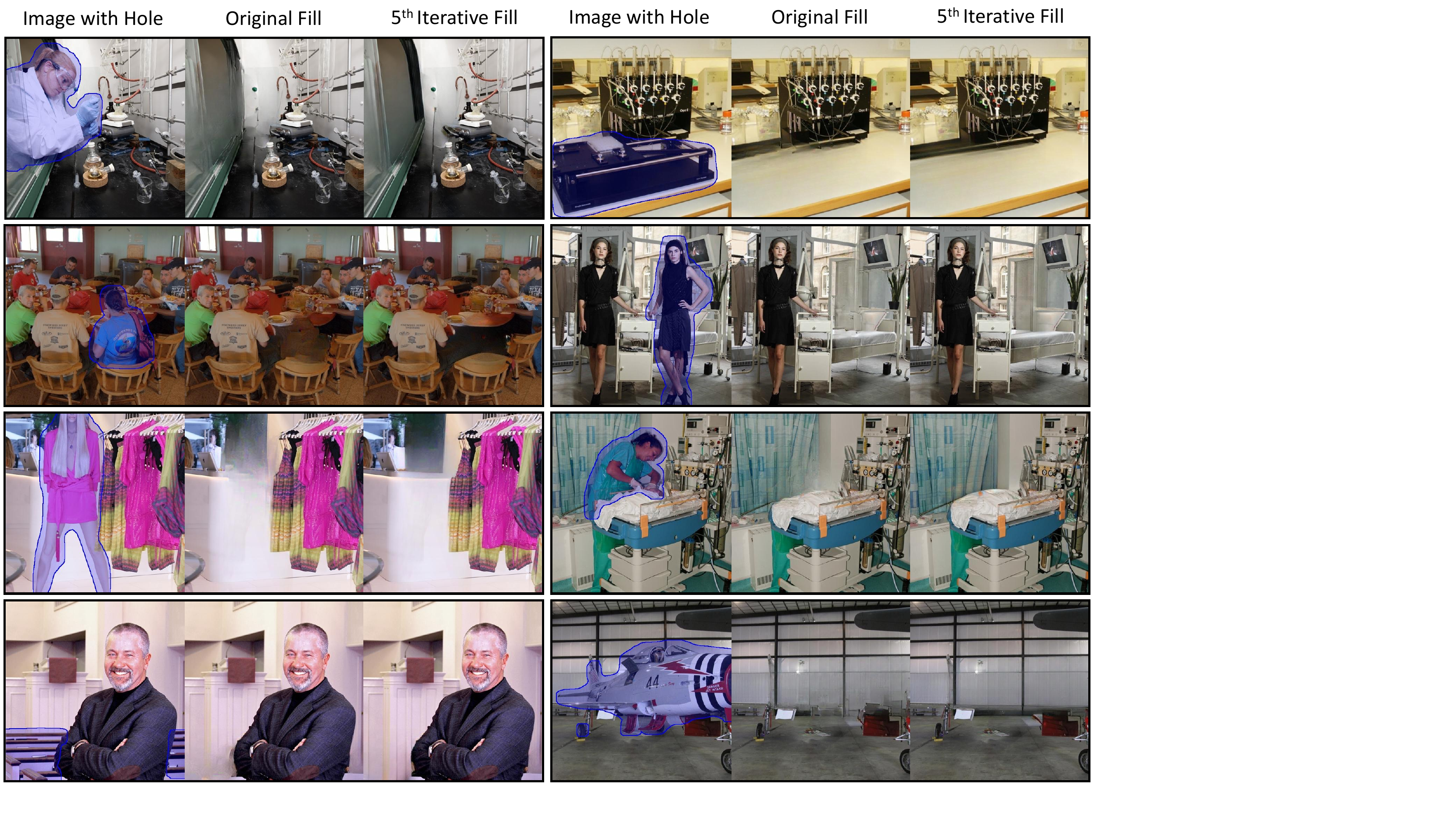}
    \vspace{-20 pt}
    \caption{Visual comparisons between the original fill and our iterative fill for LaMa \cite{suvorov2021resolution}. }
    \label{fig:lama_fill_vs_refill}
    \vspace{-5 pt}
\end{figure*}

\begin{figure*}[!h]
    \centering
    \vspace{-15 pt}
    \includegraphics[trim=0.0in 0.3in 3.2in 0.0in, clip,width=\textwidth]{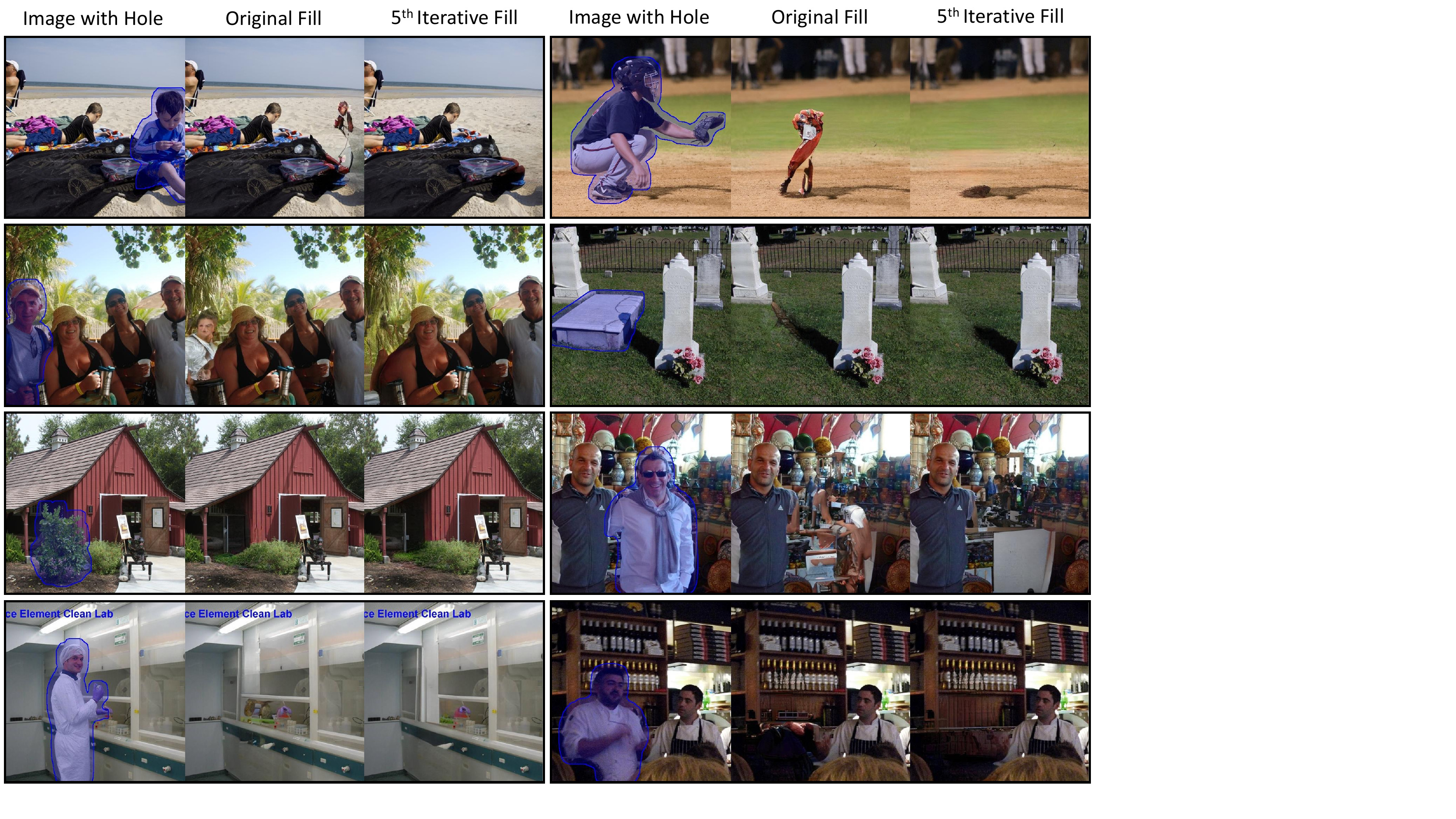}
    \vspace{-20 pt}
    \caption{Visual comparisons between the original fill and our iterative fill for CoMod-GAN \cite{zhao2021large}. }
    \label{fig:comod_fill_vs_refill}
    \vspace{-25 pt}
\end{figure*}

\begin{figure*}[!h]
    \centering
    \vspace{-10 pt}
    \includegraphics[trim=0.0in 0.3in 3.2in 0.0in, clip,width=\textwidth]{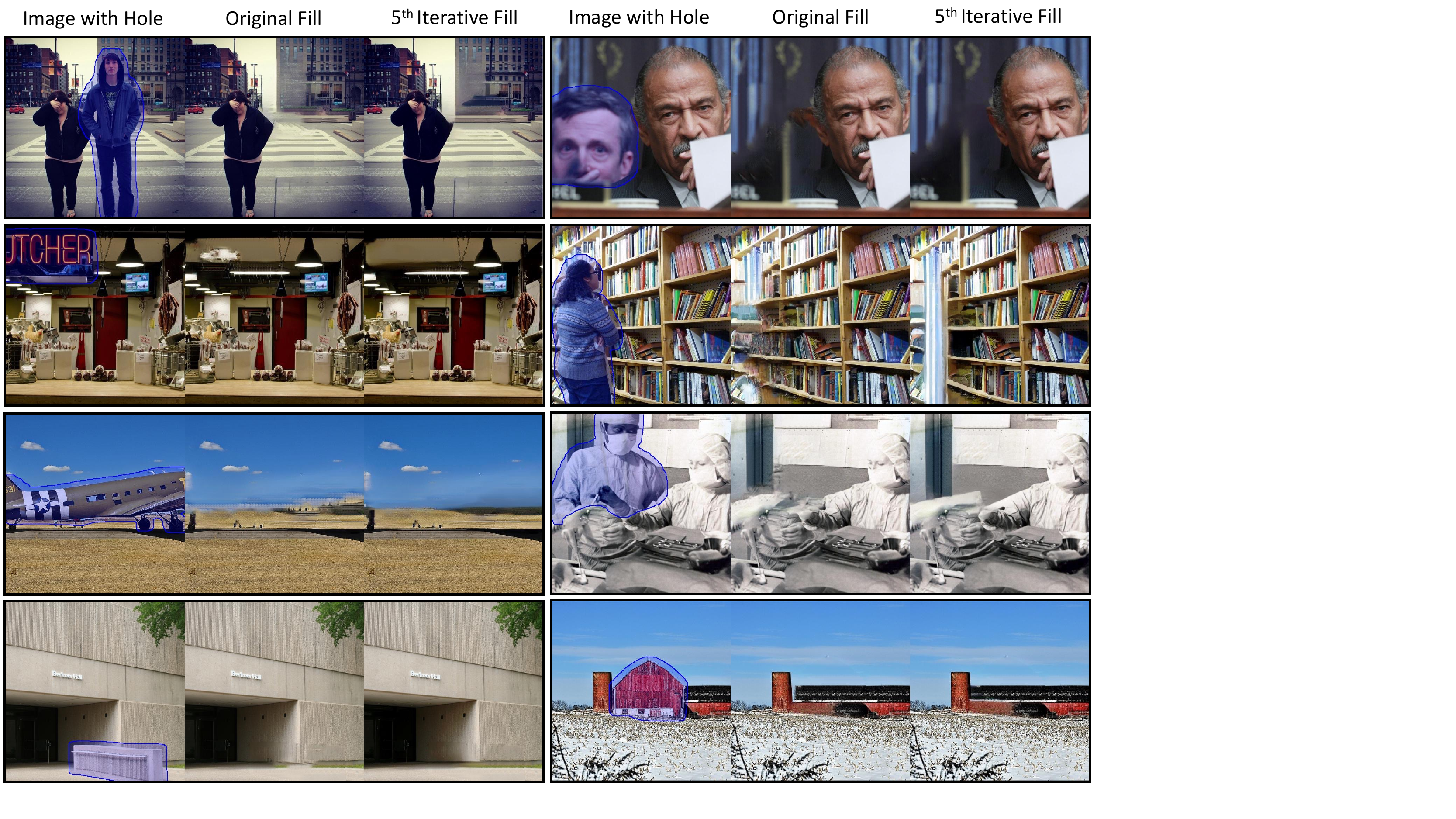}
    \vspace{-20 pt}
    \caption{Visual comparisons between the original fill and our iterative fill for ProFill \cite{zeng2020high}. }
    \label{fig:profill_fill_vs_refill}
    \vspace{-5 pt}
\end{figure*}

\begin{figure*}[!h]
    \centering
    \vspace{-10 pt}
    \includegraphics[trim=0.0in 0.3in 3.2in 0.0in, clip,width=\textwidth]{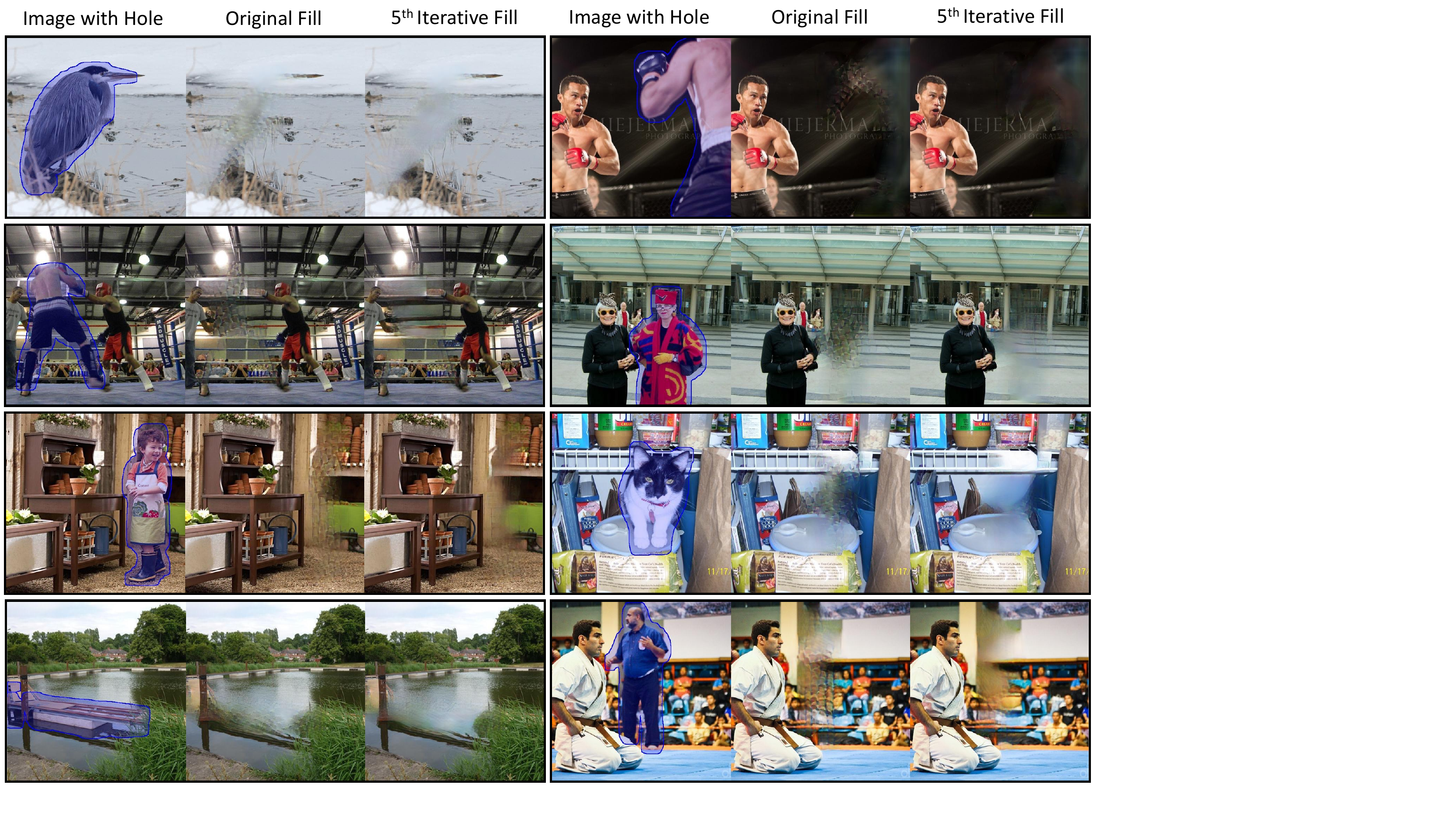}
    \vspace{-20 pt}
    \caption{Visual comparisons between the original fill and our iterative fill for EdgeConnect \cite{nazeri2019edgeconnect}. Note that the $5^{th}$ iterative fill still has obvious artifacts due to the limitation of the algorithm itself, but looks visually more pleasant than the original fill. }
    \label{fig:edgeconnect_fill_vs_refill}
    \vspace{-30 pt}
\end{figure*}

In Fig. \ref{fig:lama_fill_vs_refill}, \ref{fig:comod_fill_vs_refill}, \ref{fig:profill_fill_vs_refill}, and \ref{fig:edgeconnect_fill_vs_refill}, we show the comparisons between original fill and our iterative fill for LaMa \cite{su2020blindly}, CoMod-GAN \cite{zhao2021large}, ProFill \cite{zeng2020high}, and EdgeConnect \cite{nazeri2019edgeconnect}, respectively. 

\subsection{Situations where Iterative Fill Does Not Help}

In section 6.2 in the main paper, we have studied how many cases that users think the iteratively filled images are better, similar, or worse than the original fills. The results show that users think iterative fill and original fill are similar for lots of cases. In this section, we take a deeper look at why this is the case. 

\begin{figure*}[!h]
    \centering
    \vspace{-10pt}
    \includegraphics[trim=0.0in 0.3in 0.1in 0.0in, clip,width=\textwidth]{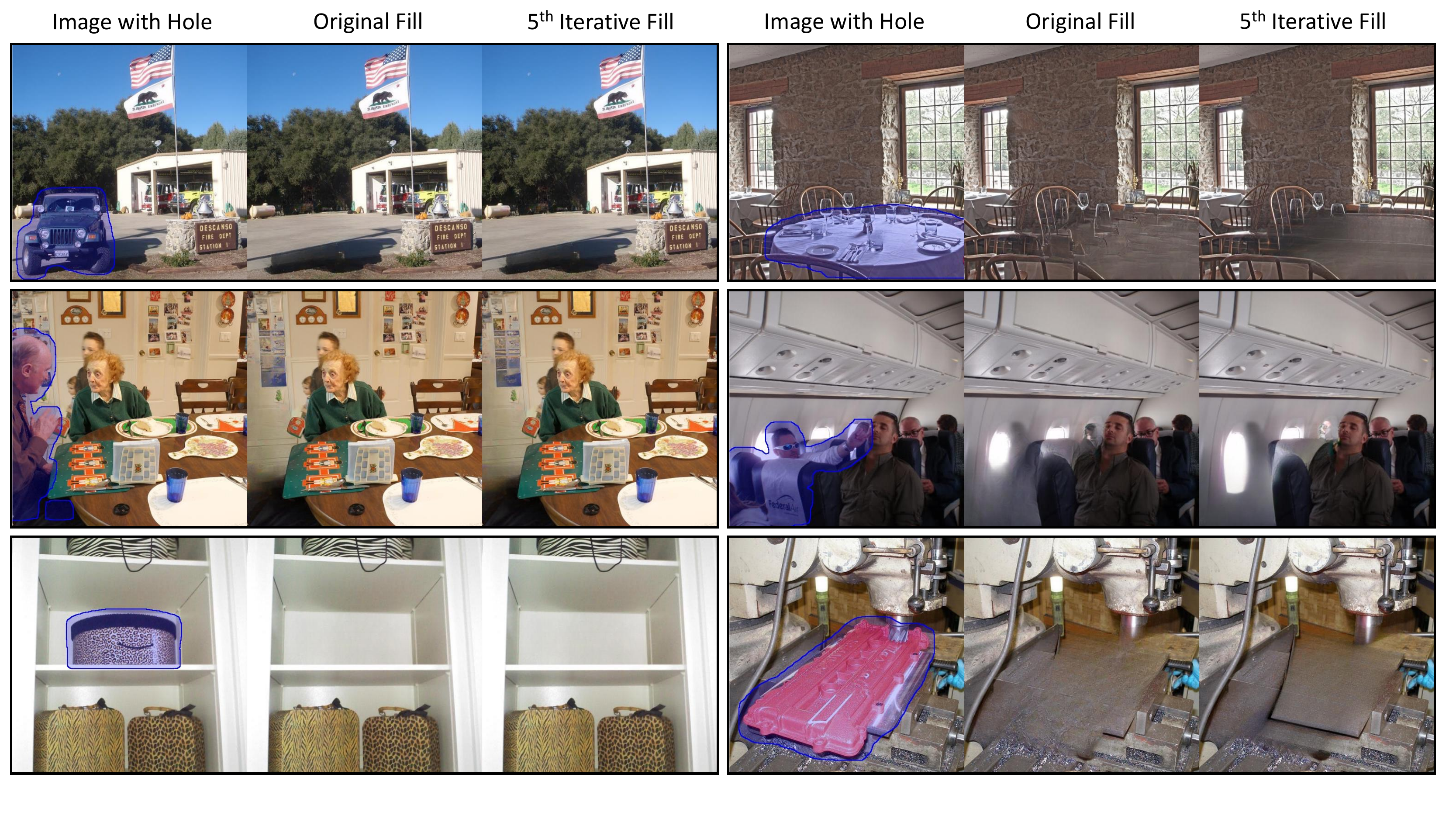}
    \vspace{-15 pt}
    \caption{Typical situations where iterative fill does not help. \textbf{Left}: when the holes are easy and original fill already looks good, our segmentation network would not detect much artifacts region and thus the iterative fill often looks very similar to the original fill. \textbf{Right}: when the holes are very large and context is limited, even if our segmentation network detects the artifacts region, iterative fill still can not properly fill the hole region and thus can not improve the inpainting quality for these hard cases.  }
    \label{fig:iterative_fill_not_help}
    \vspace{-25 pt}
\end{figure*}

In general, we observe two major reasons that cause iterative fill are similar to the original fill. First, when the holes are easy, the inpainting algorithm could already fill the holes reasonably well in a single pass. In this case, our segmentation network usually would not detect much artifacts region, and thus the iterative fill would produce very similar or even identical images as the original fill, as shown in the left of Fig. \ref{fig:iterative_fill_not_help}. Second, when the holes are large and there are not enough useful context, the original fill would usually fail obviously and our artifacts segmentation network could pick up large artifacts regions, or even as large as nearly the entire hole. However, since the useful context is still very limited, the same struggle remains almost unchanged for the same inpainting algorithm and thus iterative fill would fail to produce better outputs, and thus can not further improve the inpainting quality, as shown in the right of Fig. \ref{fig:iterative_fill_not_help}. 

%
%

\clearpage
%
%
\bibliographystyle{splncs04}
\bibliography{egbib}
\end{document}